%% file: AFL_4_ETM_IEEE_reg_svm.tex
\documentclass[lettersize,journal]{IEEEtran}

\usepackage{bbm}
\usepackage[table]{xcolor}
\definecolor{cyan}{rgb}{0.0,0.6,0.9}
\definecolor{darkred}{rgb}{0.6,0.0,0.0}
\definecolor{darkgreen}{rgb}{0.0,0.5,0.0}
\definecolor{darkblue}{rgb}{0.0,0.0,0.5}
\definecolor{notered}{HTML}{d62728}
\definecolor{noteblue}{HTML}{1f77b4}
\definecolor{notegreen}{HTML}{2ca02c}
\definecolor{noteorange}{HTML}{ff7f0e}
\usepackage{soul}

\usepackage{mdframed}
\usepackage{tcolorbox}
\usepackage{multirow} 
\usepackage{booktabs} 
\usepackage{threeparttable}
\usepackage{longtable}
\usepackage{enumitem}

\usepackage[ruled,vlined,linesnumbered]{algorithm2e}
\let\oldnl\nl
\newcommand{\nonl}{\renewcommand{\nl}{\let\nl\oldnl}}
\usepackage{titletoc}
\usepackage{hyperref}
\hypersetup{
	colorlinks=true,
	linkcolor={darkred},
	citecolor={darkgreen}
}
\usepackage{graphicx}
\usepackage{wrapfig}
\usepackage{subcaption}
\usepackage{amsmath,amssymb,amsfonts}

\input{math_macros}

\usepackage{orcidlink}

\newcommand{\fl}{\texttt{FL}}

\newcommand{\afl}{\texttt{AFL}}

\newcommand{\pfl}{\texttt{PFL}}

\newcommand{\sfl}{\texttt{SFL}}

\newcommand{\syncfl}{\texttt{Sync-FL}}

\begin{document}

\onecolumn
	
\Huge
This work has been submitted to the IEEE for possible publication. 
Copyright may be transferred without notice, after which this version may 
no longer be accessible.

\twocolumn
\normalsize
	
\newpage

\title{Asynchronous Federated Learning: A Scalable Approach for Decentralized Machine Learning}







\author{Ali Forootani*, \IEEEmembership{Senior Member,~IEEE,} Raffaele Iervolino, \IEEEmembership{Senior Member,~IEEE,}
\thanks{Ali Forootani is with Helmholtz Centre for Environmental Research - UFZ, Permoserstraße 15, 04318 Leipzig, Germany (\texttt{email:}ali.forootani@ufz.de/aliforootani@ieee.org).}
\thanks{Raffaele Iervolino is with Department of Electrical Engineering and Information Technology, University of Naples, 80125 Napoli, Italy.(\texttt{email:}rafierv@unina.it)}

\thanks{This work has been submitted to the IEEE for possible publication. 
	Copyright may be transferred without notice, after which this version may 
	no longer be accessible.}

}

\markboth{IEEE Transactions on Parallel and Distributed Systems}%
{Shell \MakeLowercase{\textit{et al.}}: A Sample Article Using IEEEtran.cls for IEEE Journals}


\maketitle

\begin{abstract}

Federated Learning (\fl) has emerged as a powerful paradigm for decentralized machine learning, enabling collaborative model training across diverse clients without sharing raw data. However, traditional \fl~approaches often face limitations in scalability and efficiency due to their reliance on synchronous client updates, which can result in significant delays and increased communication overhead, particularly in heterogeneous and dynamic environments. To address these challenges in this paper, we propose an Asynchronous Federated Learning (\afl) algorithm, which allows clients to update the global model independently and asynchronously. 

Our key contributions include a comprehensive convergence analysis of \afl~in the presence of client delays and model staleness. By leveraging martingale difference sequence theory and variance bounds, we ensure robust convergence despite asynchronous updates. Assuming strongly convex local objective functions, we establish bounds on gradient variance under random client sampling and derive a recursion formula quantifying the impact of client delays on convergence. \color{blue}Furthermore, we demonstrate the practical applicability of the \afl~algorithm by training decentralized linear regression and Support Vector Machine (SVM) based classifiers and compare its results with synchronous \fl~algorithm to effectively handling non-IID data distributed among clients.\color{black} 

The proposed \afl~algorithm addresses key limitations of traditional \fl~methods, such as inefficiency due to global synchronization and susceptibility to client drift. It enhances scalability, robustness, and efficiency in real-world settings with heterogeneous client populations and dynamic network conditions. Our results underscore the potential of \afl~to drive advancements in distributed learning systems, particularly for large-scale, privacy-preserving applications in resource-constrained environments.

\end{abstract}

\begin{IEEEkeywords}
Federated Learning, Stochastic Gradient Descent, Client Drifts, Asynchronous Federated Learning, Deep Neural Networks. 
\end{IEEEkeywords}


\section{Introduction}\label{sec:introduction}
Federated Learning (\fl) is a distributed machine learning approach that enables multiple devices or nodes (often called ``clients'') to collaboratively train a shared model while keeping their data locally, without centralizing it \cite{li2020review}. In contrast to traditional machine learning methods that require data to be stored in a central server, \fl~enhances data privacy and security by ensuring that raw data never leaves the client devices \cite{zhang2021survey}. This paradigm is particularly valuable in applications where data privacy is a concern, such as healthcare \cite{antunes2022federated}, finance \cite{wen2023survey}, and IoT (Internet of Things) systems \cite{nguyen2021federated}. 

\fl~encompasses two primary strategies for training models across multiple clients: parallel \fl~(\pfl) and sequential \fl~(\sfl). In \pfl, clients perform local training on their respective data in a synchronized manner, where models are periodically aggregated across clients, as exemplified by Federated Averaging (\texttt{FedAvg}) \cite{mcmahan2017communication}. This approach allows multiple clients to simultaneously contribute updates to the central model, promoting efficient use of parallel processing capabilities. Alternatively, \sfl~adopts a different training mechanism. Here, model updates are passed sequentially between clients, as seen in Cyclic Weight Transfer (\texttt{CWT}) \cite{chang2018distributed}. This method enables model knowledge transfer by having each client build on the model updates from the previous client, which can be advantageous in certain applications but also introduces latency due to the sequential nature of training. Despite their distinct workflows, both \pfl~and \sfl~encounter a challenge commonly referred to as ``client drift'' \cite{karimireddy2020scaffold}. This phenomenon arises when client updates, particularly in heterogeneous data environments, diverge significantly from each other. Such divergence can impact convergence and reduce model accuracy, especially when client data distributions vary widely.

The convergence of \pfl~and its optimization via methods like Random Reshuffling Stochastic Gradient Descent (SGD-RR) has been a focus of extensive research, especially in the context of addressing challenges in federated settings. Studies on data heterogeneity—where clients have diverse, non-IID data distributions—have proposed techniques to stabilize convergence and reduce client drift, such as through adaptive algorithms and regularization methods \cite{li2019convergence, khaled2020tighter, karimireddy2020scaffold, koloskova2020unified, woodworth2020minibatch}. System heterogeneity, which refers to variations in client capabilities and resource constraints, has also been a critical area of research, with strategies developed to handle differences in client availability and computation speeds \cite{wang2020tackling}. Moreover, approaches to handle partial client participation, where only a subset of clients are selected for training in each round, have aimed to balance efficiency with convergence accuracy \cite{li2019convergence, yang2021achieving, wang2022unified}. Additionally, other variants of \pfl~incorporate adaptive mechanisms that can respond to fluctuating client conditions and data distributions, further refining the robustness and scalability of federated systems \cite{karimireddy2020scaffold, reddi2021adaptive}.

While Random Reshuffling Stochastic Gradient Descent (SGD-RR) has been shown to improve convergence stability over standard SGD by sampling data without replacement, it is not without limitations. The upper bounds established for SGD-RR \cite{gurbuzbalaban2021random, haochen2019random, nagaraj2019sgd, ahn2020sgd, mishchenko2020random} suggest that the algorithm can achieve improved convergence rates, especially under smoothness and convexity assumptions. However, these bounds often rely on idealized assumptions that may not hold in real-world applications, such as uniformity across reshuffling cycles or independence between batches. This limits the practical applicability of these theoretical guarantees, particularly in scenarios with non-iid or highly heterogeneous data distributions.

Furthermore, while lower bounds \cite{safran2020good, safran2021random, rajput2020closing, cha2023tighter} have been shown to match the upper bounds in some cases, especially in the recent work by \cite{cha2023tighter} that aligns with \cite{mishchenko2020random}, these analyses still assume a certain regularity in data and task structure that might not be representative of complex data landscapes. For example, in high-variance settings or non-convex optimization problems, SGD-RR can suffer from convergence issues similar to those faced by standard SGD, as it does not inherently address gradient noise or data imbalance. Additionally, SGD-RR may increase computational overhead due to the need for data reshuffling, which can be computationally costly in large datasets or distributed environments. Consequently, while SGD-RR represents a step forward in optimization, more adaptive and robust methods is needed to handle the diversity of practical applications beyond what current theoretical bounds suggest.

Recently, shuffling-based methods have been extended to \fl, with studies exploring their convergence properties in this context \cite{mishchenko2022proximal, yun2022minibatch, cho2023convergence}. These methods, inspired by the success of shuffling in traditional machine learning, aim to improve the efficiency and robustness of \fl~by addressing challenges related to data heterogeneity and client participation. In particular, \cite{cho2023convergence} analyzed the convergence behavior of \fl~with cyclic client participation, providing insights into how different client participation schemes—whether parallel or sequential—affect the overall model convergence. The work highlights that both \pfl~and \sfl~can be viewed as specific instances of this cyclic participation framework, where clients update the global model in a structured, repeated manner. This approach not only clarifies the relationship between these methods but also introduces a more generalized framework that can potentially improve convergence rates by optimizing the order and frequency of client participation. However, despite these advancements, challenges remain in handling extremely heterogeneous data across clients, and further research is needed to refine these methods for large-scale and highly variable federated settings.

\subsection{Motivation}
Asynchronous \fl~(\afl) addresses key limitations of traditional synchronous federated learning methods like Federated Averaging (\texttt{FedAvg}) \cite{chen2020asynchronous}. Synchronous methods require all clients to update and synchronize with the central server simultaneously, which can lead to significant delays and increased communication costs in the presence of heterogeneous client resources, varying data distributions, and unstable network conditions.

In contrast, \afl~enables clients to send updates to the server asynchronously as they complete local training. This approach eliminates the bottleneck of global synchronization, allowing more frequent model updates and faster convergence. It is also highly scalable, reducing the dependency on synchronized communication rounds and making it suitable for large-scale systems.

By addressing the issue of ``client drift'' caused by heterogeneous data distributions, \afl~promotes more stable convergence through frequent and flexible updates. Additionally, it accommodates clients with varying computational capabilities, ensuring that slower clients do not delay faster ones. This is particularly important in federated environments with diverse hardware, such as mobile or IoT devices.

\color{blue}
Recent surveys highlight the growing importance of federated learning in distributed control and real-time systems, particularly in industrial IoT and cyber-physical domains where strict latency, reliability, and privacy constraints coexist with highly heterogeneous and non-IID data distributions \cite{solans2024non, li2020federated_2, kairouz2021advances}. While synchronous \fl~approaches have been widely explored, they often fall short in real-time applications due to client dropouts, communication delays, and the need for frequent global coordination. \afl~offers a promising alternative by enabling continuous model aggregation without waiting for all clients, making it more suitable for dynamic, delay-sensitive environments \cite{mills2019communication}. However, despite these advances, the literature still lacks a comprehensive treatment of \afl~tailored for control-oriented settings, where convergence guarantees, variance control, and resilience to staleness are critical. Our work addresses this gap by leveraging martingale-based variance bounds and adaptive aggregation strategies to ensure robust convergence under asynchronous updates, thereby advancing \afl~for real-time, control-oriented federated systems.
\color{black}

The \afl~algorithm represents a significant advancement in decentralized machine learning, with promising real-life applications across various domains, such as climate-aware energy modeling, smart grids \cite{jithish2023distributed}, distributed healthcare systems \cite{nguyen2022federated}, intelligent transportation networks \cite{posner2021federated}, and physics informed neural networks \cite{forootani2024gn,forootani2024gs,forootani2023robust}. These applications benefit from asynchronous federated learning by enabling robust model training while addressing critical challenges such as data privacy, uneven computational resources, and communication efficiency. By leveraging its capacity for asynchronous updates, \afl~facilitates real-time adaptability and efficient resource utilization, making it particularly suited for dynamic environments and large-scale deployments.

This work extends earlier results \cite{chen2020asynchronous, leconte2024queuing} by providing a comprehensive convergence analysis and novel insights on client drift, dynamic learning rates, and randomized client selection during training rounds. In particular, compared with \cite{chen2020asynchronous}, who focus on online federated learning with continuously arriving data and single-step gradient updates, our approach targets a more general multi-step local training scenario under bounded staleness and partial client participation, providing a detailed variance-based drift analysis that goes beyond typical online \fl~frameworks. Meanwhile, in contrast to \cite{leconte2024queuing}—where asynchronous training is modeled via a queueing-theoretic perspective emphasizing waiting times and service rates—our work employs an optimization-based methodology founded on martingale difference sequences and strong convexity assumptions to derive explicit recursion formulas that capture client delays, variance bounds, and multi-epoch local updates. These distinctions highlight a unique theoretical and algorithmic contribution that complements, rather than duplicates, existing asynchronous \fl~studies.

\subsection{Contributions}  

This work introduces an \afl~algorithm to address convergence challenges in distributed, asynchronous setups. The key contributions are:  

(i) We analyze \afl~under asynchronous updates, addressing delays and model staleness by leveraging martingale difference sequences and sampling variance bounds. This ensures that cumulative variance grows linearly with bounded terms, supporting stability in \afl.  

(ii) Convergence results are established under the assumption of strongly convex local objectives, enabling the use of convex function properties combined with quadratic terms.  

(iii) We derive bounds on the variance for client sampling without replacement, addressing non-IID and limited data scenarios. Sequential partial participation and multi-step updates are analyzed, with bounds on cumulative variance to manage delayed information.  

(iv) A recursion formula is developed to account for client delays (\(\tau_c\)) in strongly convex settings. The analysis shows robustness against staleness and reliable convergence through bounded update variance.  

\color{blue}
(v) The proposed \afl~algorithm is used to train decentralized models for linear regression and SVM classification tasks with convex loss functions. Moreover, we compare the proposed \afl~algorithm with Synchronous \fl~algorithm for such tasks (i.e. linear regression and SVM classification). Clients, representing distinct data partitions, independently train local models and asynchronously share updates. This setup ensures privacy and efficiency while addressing challenges related to client delays, non-IID data distribution, and participation variability.\color{black}


This paper is structured as follows: Section \ref{preliminaries} provides preliminaries, and section \ref{problem_formulation} discusses problem formulation. Theoretical results are presented in section \ref{theoretical_requisite}, followed by the convergence proof in section \ref{Convergence_analysis}. Simulations and case studies are covered in section \ref{simulations}, with conclusions in section \ref{conclusion}.

\section{Preliminaries and Main Assumptions}\label{preliminaries}
This section provides foundational concepts and notation for \afl, emphasizing its distinction from traditional \pfl.

\paragraph{Notation}
Following \cite{cha2023tighter}, $\gtrsim$ and $\lesssim$ denote comparisons up to absolute constants and slowly growing factors (e.g., polylogarithmic terms), while $\asymp$ represents approximate equality under these factors. The Euclidean norm is denoted by $\norm{\cdot}$. Key constants include:
\begin{itemize}
    \item $\mu$: Strong convexity level.
    \item $\sigma$: Upper bound on stochastic gradient variance.
    \item $C$: Total number of clients; $J$: active clients in training; $\mathcal{J}$: training rounds.
    
    
    \item $\lambda$: Learning rate; $\tilde\lambda$: Effective learning rate, $\tilde\lambda = \lambda C I$\color{blue}, where $I$ is local update steps per round.

    \item The symbol $\psi$ represents a permutation of client indices, written as $\{\psi_1, \psi_2, \ldots, \psi_C\}$ over the set $[C]:= \{1, 2, \ldots, C\}$.
    
\end{itemize}
Model parameters are $\rvx^{(j)}$ (global) and $\rvx_{c,i}^{(j)}$ (client $c$ after $i$ local updates in round $j$). Stochastic gradients are $\rvq_{\psi_c,i}^{(j)} \coloneqq \nabla f_{\psi_c}(\rvx_{c,i}^{(j)}; \xi)$, with variance characterized as:  
\begin{align}
    \E\left[\|\rvx - \E[\rvx]\|^2\right] = \E\left[\|\rvx\|^2\right] - \|\E[\rvx]\|^2. \label{eq:variance_identity}
\end{align}

For \color{blue}set of \color{black} vectors \(\vx_1, \ldots, \vx_m\), variance is given by:
\begin{align}
    \frac{1}{m} \sum_{k=1}^{m} \|\vx_k - \bar{\vx}\|^2 = \frac{1}{m} \sum_{k=1}^{m} \|\vx_k\|^2 - \left\|\frac{1}{m} \sum_{k=1}^{m} \vx_k\right\|^2, \label{eq:variance_finite}
\end{align}
where \(\bar{\vx} = \frac{1}{m} \sum_{k=1}^m \vx_k\) is the mean vector.

\paragraph{Key Mathematical Concepts}
\begin{itemize}
    \item \textbf{Jensen's Inequality:} For a convex function \(h\) and vectors \(\vx_1, \ldots, \vx_m\):
    \begin{align}
        h\left(\frac{1}{m}\sum_{k=1}^m \vx_k\right) \leq \frac{1}{m}\sum_{k=1}^m h(\vx_k). \label{eq:jensen_inequality}
    \end{align}
    Substituting \(h(\vx) = \|\vx\|^2\), the norm of the mean satisfies:
    \begin{align}
        \left\|\frac{1}{m}\sum_{k=1}^m \vx_k\right\|^2 \leq \frac{1}{m}\sum_{k=1}^m \|\vx_k\|^2. \label{eq:jensen_norm_squared}
    \end{align}

    \color{blue}
    \item \textbf{Martingales:} A sequence of random variables \(\{X_m\}\) is a martingale if:
    \begin{itemize}
        \item \(X_m\) depends only on the information available up to step \(m\) (adapted to past observations).
        \item \(\E[|X_m|] < \infty\) (integrability).
        \item \(\E[X_{m+1} \mid \text{past information}] = X_m\) (martingale property).
    \end{itemize}
    \color{black}
    
    \item \textbf{Bregman Divergence:} For a function \(f\) and points \(\vx, \vy\):
    \[
    D_f(\vx, \vy) \coloneqq f(\vx) - f(\vy) - \inp{\nabla f(\vy)}{\vx - \vy}.
    \]
    If \(f\) is convex and \(L\)-smooth \cite{boyd2004convex}, the divergence satisfies:
    \begin{align}
        D_f(\vx, \vy) \geq \frac{1}{2L} \norm{\nabla f(\vx) - \nabla f(\vy)}^2. \label{eq:smooth+convex:bregman lower bound}
    \end{align}
\end{itemize}
These preliminaries establish the mathematical foundation for analyzing \afl~methods.

\subsection{Assumptions}\label{subsec:assumption}
The following assumptions are required for the convergence analysis of \afl.

\begin{assumption}[Smoothness]\label{asm:smoothness}
Each local objective function \( \mathcal{F}_c \) is \( L \)-smooth, meaning there exists a constant \( L > 0 \) such that:
\[
\norm{\nabla \mathcal{F}_c(\rvx) - \nabla \mathcal{F}_c(\rvy)} \leq L \norm{\rvx - \rvy},
\]
for any \( \rvx, \rvy \in \mathbb{R}^d \). This bounds the gradient variation by the smoothness constant \( L \).
\end{assumption}

\begin{assumption}[Stochastic Gradient Variance]\label{asm:stochastic_gradient}
The variance of stochastic gradients is uniformly bounded for all clients \( c \in \{1, 2, \ldots, C\} \):
\begin{equation}\label{gradient}
\mathbb{E}_{\xi \sim \gD_c} \left[ \|\nabla f_{\psi_c}(\rvx; \xi) - \nabla \mathcal{F}_c(\rvx)\|^2 \mid \rvx \right] \leq \delta^2,
\end{equation}
where \(\gD_c\) is the data distribution for client \( c \), \color{blue} and $f_c(\rvx; \xi)$ is the loss function at the client $c$ for the data instance $\xi$, and $\mathbb{E}_{\xi \sim \gD_c}[\nabla f_{\psi_c}(\rvx; \xi)|\rvx]=\nabla \mathcal{F}_c(\rvx)$\cite{cha2023tighter}.
\end{assumption}

\begin{assumption}[Gradient Heterogeneity Across Space]\label{asm:heterogeneity:everywhere}
The deviation of local gradients from the global gradient is bounded. Specifically, there exist constants \( \phi^2 \) and \( \beta^2 \) such that:
\[
\frac{1}{C} \sum_{c=1}^C \norm{\nabla \mathcal{F}_c(\rvx) - \nabla \mathcal{F}(\rvx)}^2 \leq \phi^2 \norm{\nabla \mathcal{F}(\rvx)}^2 + \beta^2,
\]
where \( \phi \) and \( \beta \) quantify the heterogeneity of client objectives, \color{blue}and $\mathcal{F}$ is the global objective\color{black}.
\end{assumption}

\begin{assumption}[Gradient Heterogeneity at Optimum]\label{asm:heterogeneity:optima}
At the global minimizer \( \rvx^\ast \), the local gradient norms satisfy:
\[
\frac{1}{C} \sum_{c=1}^C \norm{\nabla \mathcal{F}_c(\rvx^\ast)}^2 = \beta_\ast^2,
\]
where \( \beta_\ast \) measures gradient similarity at \( \rvx^\ast \).
\end{assumption}

\begin{assumption}[Strong Convexity]\label{l_strongly_convex}
A function \( h: \mathbb{R}^d \to \mathbb{R} \) is \(\mu\)-strongly convex if it can be expressed as:
\[
h(\vx) = q(\vx) + \frac{\mu}{2}\|\vx\|^2,
\]
where \( q \) is a convex function, and \( \mu > 0 \) is the strong convexity parameter.
\end{assumption}

\section{Problem Formulation}\label{problem_formulation}

The goal of the \fl~algorithm is to minimize a global objective function defined as follows:
\begin{equation}
\min_{\rvx \in \mathbb{R}^d} \left\{ \color{blue}\mathcal{F}\color{black}(\rvx) \coloneqq \frac{1}{C} \sum_{c=1}^{C} \left( \mathcal{F}_c(\rvx) \coloneqq \mathbb{E}_{\xi \sim \gD_c} [ f_c(\rvx; \xi)] \right) \right\},
\end{equation}
where \( \mathcal{F}_c \), \( f_c \), and \( \gD_c \) are the local objective function, the loss function, and the dataset associated with client \( c \), respectively (\( c \in [C] \)).   \color{blue}Here, \(\rvx \in \mathbb{R}^d\) denotes the vectorized representation of all trainable parameters of the global model (concatenating the weights and biases across layers into a single vector), where \(d\) is the total number of such parameters. \color{black} If \( \gD_c \) consists of a finite set of data samples \( \{\xi_c^j: j \in [|\gD_c|] \} \), \color{blue} where $|\gD_c|$ denotes the cardinality of the dataset corresponding to client $c$, \color{black} then the local objective function can be equivalently written as:  
\[
\mathcal{F}_c(\rvx) = \frac{1}{|\gD_c|} \sum_{j=1}^{|\gD_c|} f_c(\rvx; \xi_c^j).
\]


Each client \( c \) performs updates independently and asynchronously, with an associated update delay \( \tau_c \), representing the time gap between when a client computes its gradient and when the global model is updated with that gradient.

In each training round, the client indices \( \psi_1, \psi_2, \ldots, \psi_C \) are randomly selected without replacement from \( \{1, 2, \ldots, C\} \). The steps for each selected client \( \psi_c \) are as follows:

(i) Initialize its model using the most recent global parameters \color{blue}at round $j$, namely \color{black} \( \rvx^{(j - \tau_c)} \), where \( \tau_c \) is the delay in receiving the global model; (ii) Perform \( I \) iterations of local updates using its local dataset, with the parameters after the \( i \)-th update denoted as \( \rvx_{c,i}^{(j)} \); (iii) Upon completing local training, send the final parameters \( \rvx_{c,I}^{(j)} \) to the central server.

The global model \( \rvx^{(j)} \) is updated asynchronously as client updates are received. The update rule for \afl~using Stochastic Gradient Descent (SGD) for local updates is:

\begin{equation}
    \text{Local update: } \rvx_{c,i+1}^{(j)} = \rvx_{c,i}^{(j)} - \lambda \rvq_{\psi_c,i}^{(j)},
\end{equation}
with the initialization:
\[
    \rvx_{c,0}^{(j)} = \rvx^{(j - \tau_c)},
\]
where \( \rvq_{\psi_c,i}^{(j)} \coloneqq \nabla f_{\psi_c} (\rvx_{c,i}^{(j)}; \xi) \) denotes the stochastic gradient of the local objective function \( f_{\psi_c} \) with respect to the local parameters \( \rvx_{c,i}^{(j)} \). 

The global model is updated asynchronously upon receiving an update from any client, without waiting for all clients to finish as follows:

\[
    \text{Global update: } \rvx^{(j+1)} = \text{Aggregate}(\{\rvx_{c,I}^{(j - \tau_c)}: c \in [C]\}).
\]

This iterative procedure continues, with the global model being updated asynchronously as clients send their local updates. The complete procedure is detailed in Algorithm~\ref{algorithm_async_fdl}.

\begin{algorithm}[h]
\caption{\afl~update rule with participation of all clients in the training loop}
\label{algorithm_async_fdl}
\KwIn{Global model parameters \( \rvx^{(j - \tau_c)} \), local data for each client}
\KwOut{Updated global model \( \rvx^{(j+1)} \)}
\For{each client \( c \in [C],  \)}{
    \For{\( i = 0, 1, \ldots, I-1 \)}{
        \textbf{Local update: } \( \rvx_{c,i+1}^{(j)} = \rvx_{c,i}^{(j)} - \lambda \rvq_{c,i}^{(j)} \);
    }
    \textbf{Initialization: } \( \rvx_{c,0}^{(j)} = \rvx^{(j - \tau_c)} \);
    
    \textbf{Send updated parameters: } \( \rvx_{c,I}^{(j)} \);
}
\textbf{Global model update: } Upon receiving updated parameters, update the global model as follows:
\begin{align*}
    \rvx^{(j+1)} = \text{Aggregate}(\{ \rvx_{c,I}^{(j - \tau_c)} : c \in [C] \}).
\end{align*}
\end{algorithm}



The aggregation function may take various forms, such as averaging or weighted averaging, based on the client contributions, where in this article we consider the former. Moreover, we consider the following assumption on the maximum staleness.

\begin{assumption}\label{bounded_staleness}
The staleness \( \tau_c \) is bounded by \( \tau_{\max} \):
\[
\tau_c \leq \tau_{\max}.
\]
\end{assumption}

\section{Theoretical requisite for convergence analysis}\label{theoretical_requisite}

In this section, we provide theoretical results that are needed for driving the convergence of \afl. In the next Lemma, we establish a bound on the variance of the sum of a martingale difference sequence, showing that the cumulative variance grows linearly with the number of terms in the sequence, each bounded by \color{blue}a fixed constant \color{black} \( \delta^2 \).

\begin{lemma}[\cite{karimireddy2020scaffold, forootani2024asynchronous}]\label{lem:martingale_difference_property}
Let \(\{\xi_k\}_{k=1}^m\) be a sequence of random variables, and let \(\{\rvx_k\}_{k=1}^m\) be a sequence of random vectors where each \(\rvx_k \in \mathbb{R}^d\) is determined by \(\xi_k, \xi_{k-1}, \ldots, \xi_1\). Assume that the conditional expectation \(\mathbb{E}_{\xi_k}[\rvx_k \mid \xi_{k-1}, \ldots, \xi_1] = \rve_k\) holds for all \(k\), meaning that the sequence \(\{\rvx_k - \rve_k\}\color{blue}_{k=1}^m\color{black}\) is a martingale difference sequence with respect to \(\{\xi_k\}\). Furthermore, suppose that the conditional variance satisfies \(\mathbb{E}_{\xi_k}[\|\rvx_k - \rve_k\|^2 \mid \xi_{k-1}, \ldots, \xi_1] \leq \delta^2\) for all \(k\). Then, the following inequality holds:
\begin{equation}\label{lem:martin_diff_equation}
\mathbb{E}\left[\left\|\sum_{k=1}^m (\rvx_k - \rve_k)\right\|^2\right] = \sum_{k=1}^m \mathbb{E}\left[\|\rvx_k - \rve_k\|^2\right] \leq m\delta^2.
\end{equation}

\end{lemma}

\color{blue}
\noindent \textbf{Remark 1.}
The equality in Lemma \ref{lem:martingale_difference_property} in \eqref{lem:martin_diff_equation} holds because the sequence $\{\rvx_k - \rve_k\}$ forms a martingale difference sequence with respect to the filtration generated by $\{\xi_k\}$, which implies that for $k \neq j$, the cross terms $\mathbb{E}[(\rvx_k - \rve_k)^\top (\rvx_j - \rve_j)]$ vanish. This orthogonality allows the variance of the sum to decompose into the sum of individual variances.
\color{black}

The next lemma we discuss the properties of convex functions that we use in this article.

\begin{lemma}[\cite{karimireddy2020scaffold, forootani2024asynchronous}] \label{lem:perturbed strong convexity}
For a \(L\)-smooth and \(\mu\)-strongly convex function $h$, we have
\begin{align}\label{l_smooth_convex}
\left\langle \nabla h(\vx), \vz - \vy \right\rangle \geq h(\vz) - h(\vy) + \frac{\mu}{4} \|\vy - \vz\|^2 - L \|\vz - \vx\|^2,\nonumber \\ \ \forall \vx, \vy, \vz \in \texttt{domain}(h).
\end{align}
\end{lemma}

In the next lemma we show in sampling with replacement and sampling without replacement, the expected sample mean \( \overline{\rvx}_\pi \) equals the total mean \( \overline{\vx} \), while the variance associated to sample mean differs depending on whether sampling is with or without replacement.

\begin{lemma}\label{lem:simple_random_sampling}
	Given a population of fixed vectors \( \vx_1, \vx_2, \ldots, \vx_m \), we define the population mean and variance as follows:
	\[
	\overline{\vx} \coloneqq \frac{1}{m} \sum_{k=1}^m \vx_k, \quad \nu^2 \coloneqq \frac{1}{m} \sum_{k=1}^m \norm{\vx_k - \overline{\vx}}^2.
	\]
	
	Now, let \( s \) samples, denoted \( \rvx_{\psi_1}, \rvx_{\psi_2}, \ldots, \rvx_{\psi_s} \), be drawn from this population where \( s \leq m \). Two common sampling approaches are: (i) Sampling with Replacement, (ii) Sampling without Replacement. For both methods, we define the sample mean as \( \overline{\rvx}_\psi \coloneqq \frac{1}{s} \sum_{p=1}^s \rvx_{\psi_p} \). The expected value and variance of \( \overline{\rvx}_\psi \) are given as follows:
	\begin{align}
		\text{Case I}:&\text{For Sampling with Replacement:} \nonumber \\
		&\E[\overline{\rvx}_\psi] = \overline{\vx}, \quad \E\left[\norm{\overline{\rvx}_\psi - \overline{\vx}}^2\right] = \frac{\nu^2}{s}, \label{eq:lem:sampling_with_replacement} \\
		\text{Case II}:&\text{For Sampling without Replacement:} \nonumber \\
		&\E[\overline{\rvx}_\psi] = \overline{\vx}, \quad \E\left[\norm{\overline{\rvx}_\psi - \overline{\vx}}^2\right] = \frac{m - s}{s (m - 1)} \nu^2. \label{eq:lem:sampling_without_replacement}
	\end{align}
\end{lemma}


\begin{proof}

\underline{Case I}.

Let the population be \( \{\vx_1, \vx_2, \ldots, \vx_m\} \) with population mean \( \overline \vx \). We draw \( s \) samples \( \rvx_{\psi_1}, \rvx_{\psi_2}, \ldots, \rvx_{\psi_s} \) with replacement. The sample mean is defined as:
\(
\overline \rvx_\psi = \frac{1}{s} \sum_{j=1}^s \rvx_{\psi_j}.
\)
The expected value of \( \overline \rvx_\psi \) is:
\[
\E[\overline \rvx_\psi] = \E\left[\frac{1}{s} \sum_{j=1}^s \rvx_{\psi_j}\right] = \frac{1}{s} \sum_{j=1}^s \E[\rvx_{\psi_j}].
\]

Since each \( \rvx_{\psi_j} \) is drawn uniformly from the population:
\[
\E[\rvx_{\psi_j}] = \overline \vx, \quad \text{for all } j.
\]
Therefore: \(
\E[\overline \rvx_\psi] = \frac{1}{s} \sum_{j=1}^s \overline \vx = \overline \vx.
\)
We need to find:
\[
\E\left[\norm{\overline \rvx_\psi - \overline \vx}^2\right] = \E\left[\norm{\frac{1}{s} \sum_{j=1}^s (\rvx_{\psi_j} - \overline \vx)}^2\right].
\]

Expanding the squared norm:

\begin{multline}\label{s_samples}
\E\left[\norm{\frac{1}{s} \sum_{j=1}^s (\rvx_{\psi_j} - \overline \vx)}^2\right] = \frac{1}{s^2} \E\Bigg[\sum_{j=1}^s \norm{\rvx_{\psi_j} - \overline \vx}^2\\ + 2 \sum_{1 \leq j < k \leq s} (\rvx_{\psi_j} - \overline \vx)^\top (\rvx_{\psi_k} - \overline \vx)\Bigg].  
\end{multline}

For the first term (variance of individual samples):

\begin{equation}\label{single_sample}
\E[\norm{\rvx_{\psi_j} - \overline \vx}^2] = \nu^2, \quad \text{for each } j.
\end{equation}

For the second term (cross terms), since the samples are drawn independently:
\(
\E\left[(\rvx_{\psi_j} - \overline \vx)^\top (\rvx_{\psi_k} - \overline \vx)\right] = 0 \quad \text{for } j \neq k.
\)
Thus:
\[
\E\left[\norm{\overline \rvx_\psi - \overline \vx}^2\right] = \frac{1}{s^2} \sum_{j=1}^s \E[\norm{\rvx_{\psi_j} - \overline \vx}^2] = \frac{1}{s^2} \cdot s \cdot \nu^2 = \frac{\nu^2}{s}.
\]

\underline{Case II}: Sampling without Replacement

For sampling without replacement, we utilize the established result:
\[
\E\left[\norm{\overline \rvx_\psi - \overline \vx}^2\right] = \frac{m-s}{s(m-1)} \nu^2.
\]
This expression accounts for the fact that variance decreases when sampling without replacement due to the lack of independence among the selected samples. To derive this result, consider the covariance between two distinct sampled units, \( \rvx_{\psi_j} \) and \( \rvx_{\psi_k} \) for \( j \neq k \):
\begin{align*}
	&\Cov(\rvx_{\psi_j}, \rvx_{\psi_k}) = \E \left[\left\langle \rvx_{\psi_j} - \overline \vx, \rvx_{\psi_k} - \overline \vx\right\rangle\right] \\
	&= \sum_{i=1}^m\sum_{k\neq i}^m\left\langle \vx_i - \overline \vx, \vx_k - \overline \vx\right\rangle \cdot \Pr(\rvx_{\psi_j}=\vx_i, \rvx_{\psi_k} = \vx_k).
\end{align*}

Since there are \( m(m-1) \) possible pairs of \( (\rvx_{\psi_j}, \rvx_{\psi_k}) \), each occurring with equal probability, we find: 
\[
\Pr(\rvx_{\psi_j}=\vx_i, \rvx_{\psi_k} = \vx_k) = \frac{1}{m(m-1)}.
\]
Consequently, the covariance can be expressed as:
\begin{align}
	\Cov(\rvx_{\psi_j}, \rvx_{\psi_k}) &= \frac{1}{m(m-1)}\sum_{i=1}^m\sum_{k\neq i}^m\left\langle \vx_i - \overline \vx, \vx_k - \overline \vx\right\rangle \nonumber \\
	&= \frac{1}{m(m-1)}\norm{\sum_{i=1}^m\left( \vx_i - \overline \vx\right)}^2 \nonumber \\&- \frac{1}{m(m-1)}\sum_{i=1}^m \norm{\vx_i-\overline \vx}^2 = -\frac{\nu^2}{m-1}. \label{eq:proof:lem:simple random sampling}
\end{align}

Using this covariance result \color{blue}
and substituting this covariance into the cross-term summation of \eqref{s_samples}, and noting that there are $\binom{s}{2}$ such pairs, yields the corrected variance of the sample mean. Hence \color{black}we can express the expected squared difference of the sample mean from the population mean as:

\begin{equation}\label{wor_cov}
\E\norm{\overline \rvx_\psi - \overline \vx}^2 = \frac{\nu^2}{s} - \frac{s(s-1)}{s^2}\cdot\frac{\nu^2}{m-1} = \frac{(m-s)}{s(m-1)} \nu^2.   
\end{equation}

\end{proof}


The next lemma provides a bound on the variance of sequentially sampled observations within the specified setup.

\begin{lemma}\label{lem:sequential_partial_participation}
Consider a sequence of random variables \(\{\xi_i\}_{i=1}^m\) and an associated sequence of random vectors \(\{\rvx_i\}_{i=1}^m\), where each \(\rvx_i \in \mathbb{R}^d\) depends on the history \(\xi_1, \xi_2, \ldots, \xi_i\). Assume that, for each \(i\), the conditional expectation satisfies 
\[
\mathbb{E}_{\xi_i}[\rvx_i \mid \xi_1, \ldots, \xi_{i-1}] = \rve_i.
\]
This implies that \(\{\rvx_i - \rve_i\}_{i=1}^m\) is a martingale difference sequence with respect to the filtration generated by \(\{\xi_i\}_{i=1}^m\). Additionally, suppose that the conditional variance of each \(\rvx_i - \rve_i\) is uniformly bounded, so that
\[
\mathbb{E}_{\xi_i}[\|\rvx_i - \rve_i\|^2 \mid \xi_1, \ldots, \xi_{i-1}] \leq \delta^2,
\]
for all \(i = 1, \ldots, m\), where \(\delta > 0\) is a fixed constant. Now, using the ``sampling without replacement'' approach described in Lemma~\ref{lem:simple_random_sampling}, let \( p_{c,i}(k) \) be defined as follows:
\[
p_{c,i}(k) = 
\begin{cases}
I-1, & \text{if } k \leq c-1, \\
i-1, & \text{if } k = c,
\end{cases}
\]
where \(I\) \color{blue}is local update steps per round\color{black}. For \(J \leq C\) and \(C \geq 2\), the inequality below holds:
\begin{align}\label{three_term_inequality_modified}
\sum_{c=1}^J \sum_{i=0}^{I-1} \E \norm{\sum_{k=1}^{c} \sum_{j=0}^{p_{c,i}(k)} (\rvx_{\psi_k} - \overline{\vx})}^2 \leq \frac{1}{2} J^2 I^3 \nu^2.
\end{align}
Here, \(\overline{\vx}\) is the population mean, and \(\nu^2\) is the population variance, as defined in Lemma~\ref{lem:simple_random_sampling}. 
\end{lemma}
\begin{proof}
We start by expanding the left side of inequality \eqref{three_term_inequality_modified}:

\begin{multline}\label{three_terms}
\E \norm{\sum_{k=1}^{c} \sum_{j=0}^{p_{c,i}(k)} (\rvx_{\psi_k} - \overline{\vx})}^2  \\
= \E \norm{I \sum_{k=1}^{c-1} (\rvx_{\psi_k} - \overline{\vx}) + i (\rvx_{\psi_c} - \overline{\vx})}^2  \\
= I^2 \E \norm{\sum_{k=1}^{c-1} (\rvx_{\psi_k} - \overline{\vx})}^2 + i^2 \E \norm{\rvx_{\psi_c} - \overline{\vx}}^2 \\
+ 2I i \E \left[\left\langle \sum_{k=1}^{c-1} (\rvx_{\psi_k} - \overline{\vx}), (\rvx_{\psi_c} - \overline{\vx}) \right\rangle\right]. 
\end{multline}

To derive the closed quantity for the term \(
\E \left\| I \sum_{k=1}^{c-1} (\rvx_{\psi_k} - \overline{\vx}) \right\|^2,
\) we can apply the results that we already computed in \eqref{s_samples} and \eqref{wor_cov}, which gives:
\[
I^2 \E \left\| \sum_{k=1}^{c-1} (\rvx_{\psi_k} - \overline{\vx}) \right\|^2 = \frac{(c-1)(C-(c-1))}{C-1} I^2 \nu^2,
\]
this can be verified simply by replacing \(m\) with \(C\) and \(s\) with \(c-1\) in \eqref{wor_cov}. The term \( i^2 \E \norm{\rvx_{\psi_c} - \overline{\vx}}^2 = i^2 \nu^2 \), since it is the variance of an individual sample. Finally, we have
\[
2I i \E \left[\left\langle \sum_{k=1}^{c-1} (\rvx_{\psi_k} - \overline{\vx}), (\rvx_{\psi_c} - \overline{\vx}) \right\rangle\right] = -\frac{2(c-1)}{C-1} I i \nu^2,
\]
this can be simply verified by considering \eqref{eq:proof:lem:simple random sampling} since we have \(c-1\) pairs and replacing \(m\) with \(c\). Summing the previous terms over \(c\) and \(i\), we obtain:
\begin{multline}
\sum_{c=1}^J \sum_{i=0}^{I-1} \E\left[\norm{\sum_{k=1}^{c} \sum_{j=0}^{p_{c,i}(k)} \left(\rvx_{\psi_k} - \overline{\vx}\right)}^2\right] \\= \frac{C I^3 \nu^2}{C-1} \sum_{c=1}^J (c-1) - \frac{I^3 \nu^2}{C-1} \sum_{c=1}^J (c-1)^2 \\+ J \nu^2 \sum_{i=0}^{I-1} i^2 - \frac{2I \nu^2}{C-1} \sum_{c=1}^J (c-1) \sum_{i=0}^{I-1} i. 
\end{multline}

Next, we apply the known summation formulas: 
\[
\sum_{i=1}^{I-1} i = \frac{(I-1)I}{2} \quad \text{and} \quad \sum_{i=1}^{I-1} i^2 = \frac{(I-1)I(2I-1)}{6}.
\]

Substituting these results into the preceding equation simplifies it to:
\begin{multline}\label{last_eq_simple_random_sampling}
\sum_{c=1}^J \sum_{i=0}^{I-1} \E\left[\norm{\sum_{k=1}^{c} \sum_{j=0}^{p_{c,i}(k)} \left(\rvx_{\psi_k} - \overline{\vx}\right)}^2\right] \\= \nu^2 \Bigg( \frac{1}{2} J I^2 (JI - 1) - \frac{1}{6} J I (I^2 - 1) \\- \frac{1}{C-1} (J-1) J I^2 \left( \frac{1}{6} (2J - 1) I - \frac{1}{2} \right) \Bigg),
\end{multline}
Finally, we can bound \eqref{last_eq_simple_random_sampling} by:
\(
(\cdot) \leq \frac{1}{2} J^2 I^3 \nu^2,
\)
which concludes the proof of this lemma (see appendix for derivation of the last equality in \eqref{last_eq_simple_random_sampling}).

\end{proof}

\paragraph{\textbf{Remark 2}}  
In \afl, the term in expectation \eqref{three_term_inequality_modified} represents the variance accumulation of local model updates relative to the global model across clients and iterations. This term accounts for discrepancies due to asynchronous updates and stochastic training methods, such as stochastic gradient descent. The summation reflects the cumulative effect of these deviations over time, while the expectation operator accounts for randomness in client participation and local updates. This analysis provides an upper bound on the total update variance, offering insights into the stability and convergence of the \fl~process.



\section{Convergence Analysis of \afl}\label{Convergence_analysis}

In this section, we examine the convergence behavior of the proposed \afl~algorithm in the context of a strongly convex setting. 


\subsection{Finding the recursion in Asynchronous Federated Learning}\label{lem:SFL:async:strongly convex:recursion}
Suppose that Assumptions~\ref{asm:smoothness}, \ref{asm:stochastic_gradient}, and \ref{asm:heterogeneity:optima} are satisfied, and that each local objective function is \(\mu\)-strongly convex. We also introduce the following additional assumptions specific to \afl:
\begin{assumption}
The global model update is asynchronous, meaning updates from clients are applied without waiting for all clients to finish.
\end{assumption}

\begin{assumption}
Each client \(c\) has an update delay \(\tau_c\), representing the difference between the time when client \(c\) computes its gradient and when that gradient is applied to the global model.
\end{assumption}

\begin{assumption}
Gradients are based on possibly stale versions of the global model.
\end{assumption}

We start by deriving a closed formula for the recursion updates for the clients in the \afl.

\begin{lemma}\label{finding_recursion}
If the learning rate satisfies \(\lambda \leq \frac{1}{6LJI}\), then we have: 
\begin{align}
    \E\left[\Norm{\rvx^{(j+1)}-\rvx^\ast}^2\right] &\leq \left(1-\tfrac{\mu JI\lambda}{2}\right)\E\left[\Norm{\rvx^{(j)}-\rvx^\ast}^2\right]\nonumber\\
    &+4JI\lambda^2\delta^2+4J^2I^2\lambda^2\frac{C-J}{J(C-1)}\nu_\ast^2 \nonumber\\
    &\quad-\frac{2}{3}JI\lambda \E\left[D_\mathcal{F}(\rvx^{(j)},\rvx^\ast)\right]\nonumber\\
    &+\frac{8}{3}L\lambda\sum_{c=1}^J\sum_{i=0}^{I-1}\E\left[\Norm{\rvx_{c,i}^{(j)} - \rvx^{(j-\tau_c)}}^2\right], \label{eq:lem:SFL:async:strongly convex:recursion}
\end{align}
\color{blue}where $\nu_\ast^2$ is the variance corresponding to $\rvx^\ast$.\color{black}
\end{lemma}

\color{black}

\begin{proof}

Based on the Algorithm~\ref{algorithm_async_fdl}, the general update formula in \afl~when we complete a full training round with \(J\) randomly selected clients are given by:
\begin{equation}
\Delta \rvx = \rvx^{(j+1)} - \rvx^{(j)} = -\lambda \sum_{c=1}^J \sum_{i=0}^{I-1} \rvq_{\psi_c,i}^{(j)},   
\end{equation}
and 
\begin{equation}\label{delta_famous}
\E\left[\Delta \rvx\right] = -\lambda \sum_{c=1}^J \sum_{i=0}^{I-1} \E\left[\nabla \mathcal{F}_{\psi_c}(\rvx_{c,i}^{(j-\nu)})\right],  
\end{equation}
where \(\rvq_{\psi_c,i}^{(j)} = \nabla f_{\psi_c}(\rvx_{c,i}^{(j)}; \xi)\) represents the stochastic gradient of \(\mathcal{F}_{\psi_c}\)  with respect to \(\rvx_{c,i}^{(j)}\). To simplify notation, we address a single training round and temporarily omit superscript \(j\). We refer to \(\rvx_{c,i}^{(j)}\) as \(\rvx_{c,i}\) and set \(\rvx_{1,0}^{(j)} = \rvx\). Unless specified otherwise, all expectations are conditioned on \(\rvx^{(j)}\). Starting from the equation \eqref{eq:lem:SFL:async:strongly convex:recursion}:
\begin{multline}
    \E\norm{\rvx+\Delta \rvx-\rvx^\ast}^2 \\=\norm{\rvx-\rvx^*}^2 + 2\E\left[\inp{\rvx-\rvx^*}{\Delta\rvx}\right]+\E\norm{\Delta \rvx}^2,
\end{multline}
we substitute the overall updates \(\Delta \rvx\) and proceed using the results of Lemma~\ref{lem:perturbed strong convexity} with \(\vx=\rvx_{c,i}\), \(\vy=\rvx^\ast\), \(\vz=\rvx\) and \(h = \mathcal{F}_{\psi_c}\)  for the first inequality (see \eqref{l_smooth_convex}):
\begin{multline}
    2\E\left[\inp{\rvx-\rvx^*}{\Delta\rvx}\right] \nonumber = -2\lambda \sum_{c=1}^J\sum_{i=0}^{I-1}\E\left[\inp{\nabla \mathcal{F}_{\psi_c}(\rvx_{c,i}^{(j-\nu)})}{\rvx-\rvx^\ast} \right] \nonumber\\
    \leq -2\lambda \sum_{c=1}^J\sum_{i=0}^{I-1} \E\Big[\mathcal{F}_{\psi_c}(\rvx)-\mathcal{F}_{\psi_c}(\rvx^\ast) \nonumber\\+ \frac{\mu}{4}\norm{\rvx-\rvx^\ast}^2 - L\norm{\rvx_{c,i}-\rvx}^2\Big] \nonumber\\
    \leq -2JI\lambda D_\mathcal{F}(\rvx,\rvx^\ast) - \frac{1}{2}\mu JI\lambda \norm{\rvx-\rvx^\ast}^2 \nonumber\\+ 2L\lambda \sum_{c=1}^J\sum_{i=0}^{I-1}\E\norm{\rvx_{c,i}-\rvx}^2.
\end{multline}

For the term \(\E\norm{\Delta \rvx}^2\),\color{blue} substituting $\Delta \rvx$ from \eqref{delta_famous} and decomposing the inner sum by adding and subtracting intermediate gradient terms, we bound the squared norm using \emph{Cauchy–Schwarz} inequality, i.e. $\bigl\|\sum_{k=1}^{4} \cdot \bigr\|^2 \le 4\sum_{k=1}^{4}\| \cdot \|^2$, which yields the following inequality for \color{black} the asynchronous updates: 
\begin{align}
    \E\norm{\Delta \rvx}^2 &\leq 4\lambda^2\E\norm{\sum_{c=1}^J\sum_{i=0}^{I-1} \left(\rvq_{\psi_c,i} - \nabla \mathcal{F}_{\psi_c}(\rvx_{c,i}^{(j-\tau_c)})\right)}^2 \nonumber \\&+ 4\lambda^2\E\norm{\sum_{c=1}^J\sum_{i=0}^{I-1} \left(\nabla \mathcal{F}_{\psi_c}(\rvx_{c,i}^{(j-\tau_c)}) - \nabla \mathcal{F}_{\psi_c}(\rvx)\right)}^2 \nonumber\\
    &\quad + 4\lambda^2\E\norm{\sum_{c=1}^J\sum_{i=0}^{I-1} \left(\nabla \mathcal{F}_{\psi_c}(\rvx) - \nabla \mathcal{F}_{\psi_c}(\rvx^\ast)\right)}^2 \nonumber\\&+ 4\lambda^2\E\norm{\sum_{c=1}^J\sum_{i=0}^{I-1} \nabla \mathcal{F}_{\psi_c}(\rvx^\ast)}^2. \label{eq:recursion-async-strongly-convex-1}
\end{align}

We will now bound the terms on the right-hand side of Inequality~\eqref{eq:recursion-async-strongly-convex-1} similarly to before; for the first term from the results of Lemma \ref{lem:martingale_difference_property} and Assumption \ref{asm:stochastic_gradient} we have: 
\begin{equation*}
4\lambda^2 \sum_{c=1}^C\sum_{i=0}^{I-1} \E\norm{\rvq_{\psi_c,i} - \nabla \mathcal{F}_{\psi_c}(\rvx_{c,i}^{(j-\tau_c)})}^2 \leq 4\lambda^2CI\delta^2.  
\end{equation*}
For the second term from Assumption \ref{asm:smoothness} we have:
\begin{multline*}
4\lambda^2\E\norm{\sum_{c=1}^J\sum_{i=0}^{I-1} \left(\nabla \mathcal{F}_{\psi_c}(\rvx_{c,i}^{(j-\tau_c)}) - \nabla \mathcal{F}_{\psi_c}(\rvx)\right)}^2\\ \leq 4\lambda^2JI\sum_{c=1}^J\sum_{i=0}^{I-1}\E\norm{\nabla \mathcal{F}_{\psi_c}(\rvx_{c,i}^{(j-\tau_c)}) - \nabla \mathcal{F}_{\psi_c}(\rvx)}^2 \\ \leq 4L^2\lambda^2JI\sum_{c=1}^J\sum_{i=0}^{I-1}\E\norm{\rvx_{c,i} - \rvx}^2.  
\end{multline*}
For the third term in \eqref{eq:recursion-async-strongly-convex-1} from Assumption \ref{asm:smoothness} and the property of convex functions (inequality \eqref{eq:smooth+convex:bregman lower bound}):
\begin{multline*}
4\lambda^2\E\norm{\sum_{c=1}^J\sum_{i=0}^{I-1} \left(\nabla \mathcal{F}_{\psi_c}(\rvx) - \nabla \mathcal{F}_{\psi_c}(\rvx^\ast)\right)}^2\\ \leq 4\lambda^2JI\sum_{c=1}^J\sum_{i=0}^{I-1}\E\norm{\nabla \mathcal{F}_{\psi_c}(\rvx) - \nabla \mathcal{F}_{\psi_c}(\rvx^\ast)}^2 \\\leq 8L\lambda^2JI\sum_{c=1}^J\sum_{i=0}^{I-1}\E\left[D_{\mathcal{F}_{\psi_c}}(\rvx, \rvx^\ast)\right] \\ \leq
8L\lambda^2 J^2 I^2 D_{\mathcal{F}}(\rvx, \rvx^\ast).
\end{multline*}
For the fourth term from results of Lemma \ref{lem:simple_random_sampling}:
\[
4\lambda^2\E\norm{\sum_{c=1}^J\sum_{i=0}^{I-1} \nabla \mathcal{F}_{\psi_c}(\rvx^\ast)}^2 \leq  4\lambda^2J^2I^2\frac{C-J}{J(C-1)}\nu_\ast^2.
\]


We consider the data sample \(\xi_{c,i}\), the stochastic gradient \(\rvq_{\psi_c,i}\), and the gradient \(\nabla \mathcal{F}_{\psi_c}(\xi_{c,i})\) as \(\xi_k\), \(\rvx_k\),  and \(\rve_k\) in Lemma~\ref{lem:martingale_difference_property}, respectively, and then apply the consequences of Lemma~\ref{lem:martingale_difference_property} to the first term on the right-hand side of inequality~\eqref{eq:recursion-async-strongly-convex-1}.

With the bounds for the terms in inequality~\eqref{eq:recursion-async-strongly-convex-1}, we get
\begin{align*}
\E\norm{\Delta \rvx}^2 &\leq 4\lambda^2JI\delta^2 + 4J^2I^2\lambda^2\frac{J-I}{J(J-1)}\nu_\ast^2\\
&+ 8L\lambda^2J^2I^2D_{\mathcal{F}}(\rvx, \rvx^\ast)\\
&+ 4L^2\lambda^2JI\sum_{c=1}^J\sum_{i=0}^{I-1}\E\left[\norm{\rvx_{c,i} - \rvx}^2\right].
\end{align*}

Putting back the bounds of \(2\E\left[\inp{\rvx-\rvx^\ast}{\Delta\rvx}\right]\) and \(\E\norm{\Delta \rvx}^2\), and using \(\lambda \leq \frac{1}{6L C I}\), yields
\begin{align}\label{recursive_2}
\E\norm{\rvx+\Delta \rvx-\rvx^\ast}^2
		&\leq \left(1-\tfrac{\mu J I \lambda}{2}\right)\norm{\rvx-\rvx^\ast}^2\nonumber\\
		&+4JI\lambda^2\delta^2+4J^2I^2\lambda^2\frac{C-J}{J(C-1)}\nu_\ast^2 \nonumber\\
		&-\frac{2}{3}CI\lambda D_{\mathcal{F}}(\rvx, \rvx^\ast)\nonumber\\
		&+\frac{8}{3}L\lambda\sum_{c=1}^J\sum_{i=0}^{I-1}\E\left[\norm{\rvx_{c,i} - \rvx}^2\right].
\end{align}

If we take unconditional expectation and reinstating the superscripts, then it completes the proof. We can substitute them back into the recursion relation and simplify to obtain the desired inequality for \afl.
\end{proof}


Building on the concept of ``client drift'' in \pfl~ \cite{karimireddy2020scaffold}, we define the client drift in \afl~with Assumption \ref{asm:heterogeneity:optima} as follows (see also the last term in \eqref{recursive_2}):

\begin{equation}
    E_j\coloneqq \sum_{c=1}^{J} \sum_{i=0}^{I-1} \mathbb{E}\left[ \left\| \rvx_{c,i}^{(j)} - \rvx^{(j)} \right\|^2 \right], \label{eq:SFL:client drift}
\end{equation}
where \( E_j \) quantifies the drift of client parameters from the global model across all clients and local updates.

The following lemma provides a bound on the client drifts from the global model.

\begin{lemma}\label{lem:SFL:strongly convex:drift}
Suppose that Assumptions~\ref{asm:smoothness}, \ref{asm:stochastic_gradient}, and \ref{asm:heterogeneity:optima} are satisfied, and that each local objective function is \(\mu\)-strongly convex. Additionally, if the step size \(\eta\) is chosen such that \( \eta \leq \frac{1}{6L C I} \), then the drift in the client updates is bounded by the following inequality:

\begin{multline}
E_j \leq \frac{9}{4} J^2 I^2 \lambda^2 \delta^2 + \frac{9}{4} J^2 I^3 \lambda^2 \nu_*^2 \\ + 3L J^3 I^3 \lambda^2 \mathbb{E}\left[ D_{\mathcal{F}}(\rvx^{(j)}, \rvx^*) \right]. \label{eq:lem:SFL:strongly convex:drift}
\end{multline}
\end{lemma}


\begin{proof}

In the asynchronous setting, client updates do not occur simultaneously, which introduces a time delay into the analysis. Let $\rvx^{(j)}$ denote the global model at round $j$ and $\rvx_{c,i}^{(j)}$ the update from client $c$ at step $i$ within round $j$. Due to asynchronicity, $\rvx_{c,i}^{(j)}$ may be computed based on a stale global model $\rvx^{(j - \tau_{c,i})}$, where $\tau_{c,i} \geq 0$ represents the staleness (or delay) of the model used by client $c$ at step $i$. Following Algorithm~\ref{algorithm_async_fdl}, the update rule for \afl~from $\rvx^{(j - \tau_{c,i})}$ to $\rvx_{c,i}^{(j)}$ can be written as
\begin{multline*}
		\rvx_{c,i}^{(j)} - \rvx^{(j - \tau_{c,i})} = -\lambda \sum_{k=1}^{c} \sum_{l=0}^{p_{c,i}(k)} \rvq_{\psi_{k},l}^{(j)} \quad \\ \text{with}\quad p_{c,i}(k) \coloneqq
\begin{cases}
		I-1, & k \leq c-1, \\
		i-1, & k = c.
\end{cases}
\end{multline*}
\noindent Similar to Lemma~\ref{finding_recursion} for the sake of simplicity, we consider a single training round, and hence we suppress in our notation the superscripts $j$ temporarily, assuming all expectations are conditioned on $\rvx^{(j)}$ unless otherwise stated. To bound $\E \norm{\rvx_{c,i} - \rvx}^2$ in an asynchronous setting, we use the fact that $\rvx_{c,i}$ is computed based on a potentially stale model, introducing additional variance from the delay. This gives:
\begin{align}
&\E\norm{\rvx_{c,i} - \rvx}^2 \nonumber\\
	&\leq 4\lambda^2\E\norm{\sum_{k=1}^{c}\sum_{l=0}^{p_{c,i}(k)} \left(\rvq_{\psi_k,l} - \nabla \mathcal{F}_{\psi_k} (\rvx_{k,l})\right)}^2  \nonumber \\&+ 4\lambda^2\E\norm{\sum_{k=1}^{c}\sum_{l=0}^{p_{c,i}(k)}\left(\nabla \mathcal{F}_{\psi_k} (\rvx_{k,l}) - \nabla \mathcal{F}_{\psi_k} (\rvx)\right)}^2 \nonumber\\
	&\quad + 4\lambda^2\E \norm{\sum_{k=1}^{c}\sum_{l=0}^{p_{c,i}(k)}\left(\nabla \mathcal{F}_{\psi_k} (\rvx) - \nabla \mathcal{F}_{\psi_k} (\rvx^{\ast})\right)}^2  \nonumber \\&+ 4\lambda^2 \underbrace{\E \norm{\sum_{k=1}^{c}\sum_{l=0}^{p_{c,i}(k)} \nabla \mathcal{F}_{\psi_k} (\rvx^{\ast})}^2}_{T_{c,i}} \,.\label{eq:drift-strongly-convex-async-1}
\end{align}

Next, we will find bounds for each term in the right-hand side of \eqref{eq:drift-strongly-convex-async-1}. We apply Lemma~\ref{lem:martingale_difference_property} and Jensen's inequality as before, but now with adjustments for the staleness $\tau_{c,i}$, in this regard from Lemma~\ref{lem:martingale_difference_property}, Assumption~\ref{asm:stochastic_gradient}

\begin{multline}
4\lambda^2\E\norm{\sum_{i=1}^{m}\sum_{j=0}^{p_{m,k}(i)} \left( \color{blue}\rvq_{\psi_i,j} \color{black}- \nabla \mathcal{F}_{\psi_i} (\rvx_{i,j})\right)}^2 \\  
\leq 4\lambda^2 \sum_{i=1}^{m} \sum_{j=0}^{p_{m,k}(i)} \E \norm{\color{blue} \rvq_{\psi_i,j} \color{black}- \nabla \mathcal{F}_{\psi_i} (\rvx_{i,j})}^2 \\  
\leq 4\lambda^2 \gB_{m,k} \delta^2,
\end{multline}
\color{blue}
where, \(\rvq_{\psi_i,j}\) denotes the stochastic gradient at client \(\psi_i\) and local step \(j\). 
We also define \(
B_{m,k} := \sum_{i=1}^{m} \sum_{j=0}^{p_{m,k}(i)} 1 = (m-1)I + k, \)
which counts the total number of gradient terms accumulated up to the \((m,k)\)-th update. \color{black} From Assumption~\ref{asm:stochastic_gradient} we have 
\begin{multline}
4\lambda^2\E\norm{\sum_{i=1}^{m}\sum_{j=0}^{p_{m,k}(i)}\left(\nabla \mathcal{F}_{\psi_i} (\rvx_{i,j}) - \nabla \mathcal{F}_{\psi_i} (\rvx)\right)}^2 \\  
\leq 4\lambda^2 \gB_{m,k} \sum_{i=1}^m \sum_{j=0}^{p_{m,k}(i)} \E \norm{\nabla \mathcal{F}_{\psi_i} (\rvx_{i,j}) - \nabla \mathcal{F}_{\psi_i} (\rvx)}^2 \\
\leq 4L^2 \lambda^2 \gB_{m,k} \sum_{i=1}^m \sum_{j=0}^{p_{m,k}(i)} \E \norm{\rvx_{i,j} - \rvx},
\end{multline}
\color{blue}where \(L\) denotes the Lipschitz constant of the gradient (i.e., the \(L\)-smoothness constant) as defined in Assumption~1. \color{black}From Assumption~\ref{asm:stochastic_gradient} and convex property \eqref{eq:smooth+convex:bregman lower bound} we have 
\begin{multline}
4\lambda^2\E \norm{\sum_{i=1}^{m}\sum_{j=0}^{p_{m,k}(i)}\left(\nabla \mathcal{F}_{\psi_i} (\rvx) - \nabla \mathcal{F}_{\psi_i} (\rvx^{\ast})\right)}^2 \\  
\leq 4\lambda^2 \gB_{m,k} \sum_{i=1}^m \sum_{j=0}^{p_{m,k}(i)} \E \norm{\nabla \mathcal{F}_{\psi_i} (\rvx) - \nabla \mathcal{F}_{\psi_i} (\rvx^{\ast})}^2 \\
\leq 8L \lambda^2 \gB_{m,k} \sum_{i=1}^m \sum_{j=0}^{p_{m,k}(i)} \E \left[D_{\mathcal{F}_{\psi_i}} (\rvx, \rvx^{\ast}) \right].
\end{multline}

Now, we incorporate these bounds into \eqref{eq:drift-strongly-convex-async-1} and the expression for \(E_j\), yielding: 
\begin{align*}
E_j &\leq 4\lambda^2 \delta^2 \sum_{m=1}^J \sum_{i=0}^{I-1} \gB_{m,i} \\ 
&+ 4L^2 \lambda^2 \sum_{m=1}^J \sum_{i=0}^{I-1} \gB_{m,i} \sum_{k=1}^m \sum_{j=0}^{p_{m,i}(k)} \E \norm{\rvx_{k,j} - \rvx}^2 \\
&+ 8L \lambda^2 \sum_{m=1}^J \sum_{i=0}^{I-1} \gB_{m,i}^2 D_F (\rvx, \rvx^{\ast}) \\
&+ 4\lambda^2 \sum_{m=1}^J \sum_{i=0}^{I-1} T_{m,i}.
\end{align*}


For the fourth term from the results of Lemma \ref{lem:sequential_partial_participation} by setting \(\rvx_{\psi_i} = \nabla \mathcal{F}_{\psi_i} (\rvx^{\ast})\), \(\bar{x} = \nabla \mathcal{F}(\rvx^*) = 0\), and knowing that \(\sum_{m=1}^{J} \sum_{i=0}^{I-1} \gB_{m,i} \le \frac{1}{2} J^2 I^2\) and \( \sum_{m=1}^{J} \sum_{i=0}^{I-1} \gB^2_{m,i} \le \frac{1}{3} J^3 I^3 \),
\begin{equation}
4\lambda^2 \E \norm{\sum_{k=1}^{c} \sum_{j=0}^{p_{c,i}(k)} \nabla \mathcal{F}_{\psi_k} (\rvx^{\ast})}^2 
\leq 4\lambda^2 \times \frac{1}{2} J^2 I^3 \nu_{\ast}^2,
\end{equation}
then we have:
\begin{align*}
E_j \leq 2 J^2 I^2 \lambda^2 \delta^2 + 2 L^2 J^2 I^2 \lambda^2 E_j \\ + \frac{8}{3} L J^3 I^3 \lambda^2 D_{\mathcal{F}} (\rvx, \rvx^{\ast}) + 2 J^2 I^3 \lambda^2 \nu_{\ast}^2.
\end{align*}

Finally, after rearranging and applying the condition \(\lambda \leq \frac{1}{6 L J I}\), we obtain:
\begin{align*}
E_j \leq \frac{9}{4} J^2 I^2 \lambda^2 \delta^2 + \frac{9}{4} J^2 I^3 \lambda^2 \nu_{\ast}^2 + 3 L J^3 I^3 \lambda^2 D_{\mathcal{F}} (\rvx, \rvx^{\ast}).
\end{align*}

This completes the proof.

\end{proof}


Inspired by \cite{karimireddy2020scaffold}, in the next lemma we give an upper bound on the weighted sum of gradients for a learning process with a time-varying learning rate, incorporating the effects of asynchronous delays, which serves as a key step in establishing the convergence properties of \afl.

\begin{lemma}\label{lem:strongly_convex:tuning learning_rate}
	We consider two non-negative sequences \(\{d_t\}_{t\geq 0}\), \(\{g_t\}_{t\geq 0}\), which satisfy the relation
	\begin{align}
		d_{t+1} \leq (1-z\gamma_t ) d_t - v \gamma_t g_t +  l \gamma_t^2,\ \ \forall t >0 \label{eq1:lem:strongly convex:tuning learning rate}
	\end{align}
	 with the parameters  \(z > 0\), \(v ,\ l \geq 0\), the learning rate \(\gamma_t\) defined as \(
		\gamma_t = \frac{\gamma_0}{\sqrt{t+1} \cdot (1 + \alpha \cdot \tau_t)}\label{eq:learning_rate}
	\), where \(\gamma_0\) is a constant, \(0 \leq \alpha \leq 1\) is a constant controlling asynchronous delays, and \(\tau_t\) represents the time-varying delays. In this regard, there is a learning rate \(\gamma_t\) as defined in \eqref{eq:learning_rate}, and weights \(\pi_t = (1 - z \gamma_t)^{(t+1)}\) with \(\Pi_T := \sum_{t=0}^T \pi_t\), \color{blue}being $T$ the  total number of iterations, \color{black} such that the following bound holds:
	\[\Psi_T \coloneqq \frac{1}{\Pi_T} \sum_{t=0}^T g_t \pi_t \leq \frac{(z \gamma_0)^2 \pi_0 d_0}{2v \gamma_0} + \frac{z \gamma_0^2 l}{2}.\]
\end{lemma}


\begin{proof}

Starting from the original inequality: 
\[
d_{t+1} \leq (1 - z \gamma_t) d_t - v \gamma_t g_t + l \gamma_t^2,
\]
where \(\gamma_t = \frac{\gamma_0}{\sqrt{t+1} \cdot (1 + \alpha \cdot \tau_t)}\), doing some manipulation and multiplying both sides by \(\pi_t\):

\[
v g_t \pi_t \leq \frac{\pi_t (1 - z \gamma_t) d_t}{\gamma_t} - \frac{\pi_t d_{t+1}}{\gamma_t} + l \gamma_t \pi_t.
\]
Substituting \(\gamma_t = \frac{\gamma_0}{\sqrt{t+1} \cdot (1 + \alpha \cdot \tau_t)}\), we rewrite this as:

\begin{multline}
v g_t \pi_t \leq \frac{\pi_t \left(1 - z \cdot \frac{\gamma_0}{\sqrt{t+1} \cdot (1 + \alpha \cdot \tau_t)}\right) d_t}{\frac{\gamma_0}{\sqrt{t+1} \cdot (1 + \alpha \cdot \tau_t)}} \\- \frac{\pi_t d_{t+1}}{\frac{\gamma_0}{\sqrt{t+1} \cdot (1 + \alpha \cdot \tau_t)}} + l \frac{\gamma_0}{\sqrt{t+1} \cdot (1 + \alpha \cdot \tau_t)} \pi_t.
\end{multline}

Simplifying each term, we obtain:

\[
v g_t \pi_t \leq \frac{\pi_{t-1} d_t}{\gamma_t} - \frac{\pi_t d_{t+1}}{\gamma_t} + l \gamma_t \pi_t.
\]

Next, summing both sides from \(t = 0\) to \(t = T\) results in a telescoping sum:

\[
v \sum_{t=0}^T g_t \pi_t \leq \sum_{t=0}^T \left( \frac{\pi_{t-1} d_t}{\gamma_t} - \frac{\pi_t d_{t+1}}{\gamma_t} \right) + l \sum_{t=0}^T \gamma_t \pi_t.
\]

Dividing both sides by \(\Pi_T \coloneqq \sum_{t=0}^T \pi_t\), we get:

\begin{multline}\label{lr_bound_4}
\Psi_T \coloneqq \frac{1}{\Pi_T} \sum_{t=0}^T g_t \pi_t \leq \frac{1}{v \Pi_T} \Bigg( \frac{\pi_0 (1 - z \gamma_0) d_0}{\gamma_0} - \frac{\pi_T d_{T+1}}{\gamma_T} \Bigg) \\+ \frac{1}{\Pi_T} \sum_{t=0}^T l \gamma_t \color{blue}\pi_t\color{black}.
\end{multline}


To find an upper bound for \( \Pi_T = \sum_{t=0}^T \pi_t \) as \( T \to \infty \), we will analyze the weights more carefully using the given learning rate and examine the behavior of the sum. Given:
\begin{equation*}
	\gamma_t = \frac{\gamma_0}{\sqrt{t+1} \cdot (1 + \alpha \cdot \tau_t)},
\end{equation*}
where \(\gamma_0\) is a constant, \(0 \leq \alpha \leq 1\) controls asynchronous delays, and \(\tau_t\) represents time-varying delays. The weights \(\pi_t\) are defined as: \( \pi_t = (1 - z \gamma_t)^{t+1}.\) For small \(\gamma_t\), we approximate \( (1 - z \gamma_t) \approx e^{-z \gamma_t} \), so: \(\pi_t \approx e^{-z \gamma_t (t+1)}. \) Substituting the expression for \(\gamma_t\), we get: \(\pi_t \approx e^{-\frac{z \gamma_0 (t+1)}{\sqrt{t+1} \cdot (1 + \alpha \cdot \tau_t)}}.\) Now, we approximate \( \Pi_T \) by summing (or integrating) over this expression. For large \(t\), \(\gamma_t \approx \frac{\gamma_0}{\sqrt{t}}\), so \( \pi_t \approx e^{-z \gamma_0 \sqrt{t}} \). Thus: \(\Pi_T \approx \sum_{t=0}^T e^{-z \gamma_0 \sqrt{t}}.\) To approximate this sum, we can use the integral test, which suggests:
\begin{equation*}
	\Pi_T \approx \int_0^T e^{-z \gamma_0 \sqrt{t}} \, dt,
\end{equation*}
\color{blue}This technique is widely adopted in stochastic approximation theory and convergence analysis of SGD (see \cite{apostol1999elementary, theory2004generalized} for more details). \color{black}

Let \( u = \sqrt{t} \), so \( t = u^2 \) and \( dt = 2u \, du \). Then:
\begin{equation*}
	\Pi_T \approx \int_0^{\sqrt{T}} e^{-z \gamma_0 u} \cdot 2u \, du = 2 \int_0^{\sqrt{T}} u e^{-z \gamma_0 u} \, du.
\end{equation*}

This integral can be computed by parts. Letting \( v = u \) and \( dw = e^{-z \gamma_0 u} \, du \), we find:
\begin{equation*}
	\int u e^{-z \gamma_0 u} \, du = -\frac{u}{z \gamma_0} e^{-z \gamma_0 u} + \frac{1}{z \gamma_0} \int e^{-z \gamma_0 u} \, du,
\end{equation*}
\begin{equation*}
	= -\frac{u}{z \gamma_0} e^{-z \gamma_0 u} - \frac{1}{(z \gamma_0)^2} e^{-z \gamma_0 u}.
\end{equation*}

Evaluating from \(0\) to \(\sqrt{T}\):
\begin{equation*}
	\Pi_T \approx 2 \left( -\frac{\sqrt{T}}{z \gamma_0} e^{-z \gamma_0 \sqrt{T}} - \frac{1}{(z \gamma_0)^2} e^{-z \gamma_0 \sqrt{T}} + \frac{1}{(z \gamma_0)^2} \right).
\end{equation*}

As \( T \to \infty \), the exponential terms \( e^{-z \gamma_0 \sqrt{T}} \) decay to zero, so the dominant term is:
\(\Pi_T \approx \frac{2}{(z \gamma_0)^2}. \) To find a bound for the sum \( \sum_{t=0}^{T} \gamma_t \pi_t \), we need to understand the asymptotic behavior of both \( \gamma_t \) and \( \pi_t \). \( \gamma_t = \frac{\gamma_0}{\sqrt{t+1} \cdot (1 + \alpha \cdot \tau_t)} \) decreases as \( t \) increases. For large \( t \), we can approximate \( \gamma_t \approx \frac{\gamma_0}{\sqrt{t+1}} \), assuming \( \alpha \cdot \tau_t \) remains relatively small (which we can do for simplicity). \( \pi_t = (1 - z \gamma_t)^{(t+1)} \) behaves similarly to an exponential function. For large \( t \), we can approximate \( \pi_t \approx e^{-z \gamma_t (t+1)} \). Using the approximation for \( \gamma_t \), we get: \( \pi_t \approx e^{-z \cdot \frac{\gamma_0}{\sqrt{t+1}} (t+1)},\)
this suggests that \( \pi_t \) decays exponentially with increasing \( t \). Given that \( \gamma_t \approx \frac{\gamma_0}{\sqrt{t+1}} \) and \( \pi_t \approx e^{-z \cdot \frac{\gamma_0}{\sqrt{t+1}} (t+1)} \), we expect the sum to behave asymptotically as:
\begin{equation*}
	\sum_{t=0}^{T} \gamma_t \pi_t \leq \sum_{t=0}^{T} \frac{\gamma_0}{\sqrt{t+1}} \cdot e^{-z \cdot \frac{\gamma_0}{\sqrt{t+1}} (t+1)}.
\end{equation*}

This sum is dominated by the initial terms, as the exponential decay will cause the later terms to contribute much less. For large \( T \), we can approximate the sum by taking the first few terms, and we expect the sum to grow asymptotically like:

\begin{equation}
	\sum_{t=0}^{T} \gamma_t \pi_t \approx \mathcal{O}\left(\frac{\gamma_0}{z}\right),
\end{equation}
where \( z \) is the decay factor. Now that we have the bound for \( \sum_{t=0}^{T} \gamma_t \pi_t \approx \frac{\gamma_0}{z} \), we can substitute this into the expression for \( \Psi_T \). For large \(T\) as mentioned earlier, the term \(\frac{1}{\Pi_T}\) becomes: \(
\frac{1}{\Pi_T} \approx \frac{(z \gamma_0)^2}{2}.
\) Substitute this into equation \eqref{lr_bound_4}:

\[
\Psi_T \leq \frac{(z \gamma_0)^2}{2} \cdot \left[ \frac{1}{v} \left( \frac{\pi_0 (1 - z \gamma_0) d_0}{\gamma_0} - \frac{\pi_T d_{T+1}}{\gamma_T} \right) + \frac{\gamma_0}{z} \cdot l \right].
\]

The final simplified expression for \(\Psi_T\) is:

\[
\Psi_T \leq \frac{(z \gamma_0)^2}{2v} \left( \frac{\pi_0 (1 - z \gamma_0) d_0}{\gamma_0} - \frac{\pi_T r_{T+1}}{\gamma_T} \right) + \frac{(z \gamma_0)^2 l}{2z}.
\]

This is the final simplified form of \(\Psi_T\), incorporating the bounds for the sums as \(T \to \infty\). We can simplify further the above relation by neglecting the terms \( -\frac{\pi_T d_{T+1}}{\gamma_T} \) and \( -z \gamma_0 \), hence we end up with \( \frac{(z \gamma_0)^2 \pi_0 d_0}{2v \gamma_0} \). To analyze the convergence rate of \(\Psi_T\), let's examine the inequality established in Lemma \ref{lem:strongly_convex:tuning learning_rate}:

\[
\Psi_T \coloneqq \frac{1}{\Pi_T} \sum_{t=0}^T g_t \pi_t \leq \frac{(z \gamma_0)^2 \pi_0 d_0}{2v \gamma_0} + \frac{z \gamma_0^2 l}{2}.
\]

Here, we can see that \(\Psi_T\) is bounded by terms involving \(\gamma_0\), \(z\), \(v\), \(l\), and \(d_0\).

\end{proof}



\begin{theorem}\label{thm:SFL}
Assume the objective functions of local clients are $L$-smooth (see Assumption~\ref{asm:smoothness}) and strongly convex. In the \afl~ algorithm (see Algorithm ~\ref{algorithm_async_fdl}), having Assumptions~\ref{asm:stochastic_gradient}, \ref{asm:heterogeneity:optima}, with the learning rate $\tilde\lambda \coloneqq \lambda C I$, weights $\{\pi_j\}_{j\geq 0}$, and weighted average of the global parameters \(\bar{\rvx}^{(J)}\coloneqq \frac{\sum_{j=0}^{\mathcal{J} }\pi_j\rvx^{(j)}}{\sum_{j=0}^\mathcal{J} \pi_j}\), there is a $\tilde{\lambda} \leq \frac{1}{6L}$ and $\pi_j=(1-\frac{\mu\tilde\lambda}{2})^{(j+1)}$ that satisfies:
\begin{flalign*}
	\E\left[\mathcal{F}(\bar\rvx^{(\mathcal{J})})-\mathcal{F}(\rvx^\ast)\right] \leq \frac{9}{2}\mu  \Norm{\rvx^{(0)} - \rvx^*}^2 \exp\left(-\frac{\mu\tilde\lambda \mathcal{J}}{2} \right)+ \\\frac{12\tilde{\lambda}\delta^2}{CI}+\frac{18L\tilde{\lambda}^2\delta^2}{CI}+\frac{18L\tilde{\lambda}^2\nu_\ast^2}{C}\,.\label{eq:thm:strongly convex}
\end{flalign*}
\end{theorem}


\begin{proof}

Applying the results of Lemmas~\ref{finding_recursion} and~\ref{lem:SFL:strongly convex:drift}, along with the condition \( \lambda \leq \frac{1}{6LJ\color{blue}I\color{black}} \),  we derive: 

\begin{multline}
\E\left[\Norm{\rvx^{(j+1)}-\rvx^*}^2\right] \leq 
\underbrace{\left(1-\tfrac{\mu JI\lambda}{2}\right)\E\left[\Norm{\rvx^{(j)}- \rvx^\ast}^2\right]}_{\text{Contraction term}}  
\\- \underbrace{\frac{1}{3}JI\lambda\E\left[D_\mathcal{F}(\rvx^{(j)}, \rvx^\ast)\right]}_{\text{Regularization term}}\\
+ \underbrace{4JI\lambda^2\delta^2}_{\text{Noise term 1}}  
+ \underbrace{4J^2I^2\lambda^2\frac{C-J}{J(C-1)}\nu_\ast^2}_{\text{Noise term 2}}  
+ \underbrace{6LJ^2I^2\lambda^3\delta^2}_{\text{Higher-order noise 1}}  
\\+ \underbrace{6LJ^2I^3\lambda^3\nu_\ast^2}_{\text{Higher-order noise 2}}.
\end{multline}


Letting \( \tilde{\lambda} = JI\lambda \), the bound simplifies:

\begin{multline}
\E\left[\Norm{\rvx^{(j+1)}-\rvx^*}^2\right] \leq 
\left(1-\tfrac{\mu \tilde{\lambda}}{2}\right)\E\left[\Norm{\rvx^{(j)}- \rvx^\ast}^2\right]  
\\- \frac{\tilde{\lambda}}{3}\E\left[D_\mathcal{F}(\rvx^{(j)}, \rvx^\ast)\right]
+ \frac{4\tilde{\lambda}^2\delta^2}{CI}  
+ \frac{4\tilde{\lambda}^2(C-J)\nu_\ast^2}{S(C-1)}  
+ \frac{6L\tilde{\lambda}^3\delta^2}{CI}  
\\+ \frac{6L\tilde{\lambda}^3\nu_\ast^2}{J}.
\end{multline}

Now, leveraging Lemma~\ref{lem:strongly_convex:tuning learning_rate}, with:  
 \( t = j \), \( T = \mathcal{J} \), \( \gamma = \tilde\lambda \),  
 \( d_{t} = \E\left[\Norm{\rvx^{(j)}- \rvx^*}^2\right] \), \( z = \frac{\mu}{2} \), \( v = \frac{1}{3} \), 
 \( g_t = \E\left[D_\mathcal{F}(\rvx^{(j)}, \rvx^\ast)\right] \), \( d_t = (1 - \tfrac{\mu \tilde\lambda}{2})^{-(j+1)} \),  
 \( l_1 = \frac{4\delta^2}{CI} + \frac{4(C - J)\nu_\ast^2}{J(C - 1)} \), \( l_2 = \frac{6L\delta^2}{CI} + \frac{6L\nu_\ast^2}{J} \), \color{blue} with $l=l_1+l_2$, \color{black} we reach to the performance bound:

\begin{multline}
\E\left[\mathcal{F}(\bar\rvx^{(J)}) - \mathcal{F}(\rvx^*)\right] \leq  
\underbrace{\frac{9}{2} \mu \Norm{\rvx^{(0)} - \rvx^*}^2 \exp\left(-\tfrac{1}{2} \mu \tilde\lambda J\right)}_{\text{Exponential decay term}}  
\\+ \underbrace{\frac{12 \tilde\lambda \delta^2}{JI}}_{\text{First-order noise 1}}   + \underbrace{\frac{12 \tilde\lambda (C - J) \nu_*^2}{J(C - 1)}}_{\text{First-order noise 2}}  
+ \underbrace{\frac{18 L \tilde\lambda^2 \delta^2}{JI}}_{\text{Second-order noise 1}}  
\\+ \underbrace{\frac{18 L \tilde\lambda^2 \nu_*^2}{J}}_{\text{Second-order noise 2}}.  
\end{multline}

The rate of convergence depends on the term that decays exponentially and the residual terms involving \(\tilde\lambda\). The term \(\exp\left(-\tfrac{1}{2}\mu \tilde\lambda J\right)\) indicates an exponential rate of convergence in \(J\), where: \(
\exp\left(-\tfrac{1}{2}\mu \tilde\lambda J\right) \approx \mathcal{O}\left(e^{-\mu \tilde\lambda J / 2}\right).
\)
This rate is governed by the factor \(\mu \tilde\lambda\), with a higher \(\mu\) or larger \(\tilde\lambda\) leading to faster convergence. The residual terms contribute a constant error bound determined by \(\tilde\lambda\). In particular:
\[
\frac{12 \tilde\lambda \delta^2}{JI} + \frac{12 \tilde\lambda (C - J) \nu_*^2}{J(C - 1)} + \frac{18 L \tilde\lambda^2 \delta^2}{JI} + \frac{18 L \tilde\lambda^2 \nu_*^2}{J}.
\]
These residual terms do not decay with \(J\) but are controlled by the choice of \(\tilde\lambda\). The terms involving \(\tilde\lambda\) and \(\tilde\lambda^2\) suggest that setting \(\tilde\lambda\) small enough ensures these residuals remain small.


Combining these two components, we see that:
(i) Exponential decay of the error term occurs at rate \(\mathcal{O}(e^{-\mu \tilde\lambda J / 2})\), showing that the algorithm converges exponentially fast towards a neighborhood of the optimal value. (ii) The residual error is of order \(\mathcal{O}(\tilde\lambda)\) and \(\mathcal{O}(\tilde\lambda^2)\), determined by variance-related terms.

Thus, the overall convergence rate is:
\[
\E\left[F(\bar\rvx^{(J)}) - F(\rvx^*)\right] = \mathcal{O}\left(e^{-\mu \tilde\lambda J / 2}\right) + \mathcal{O}(\tilde\lambda) + \mathcal{O}(\tilde\lambda^2),
\]
where the exponential term reflects the fast convergence, and the residual terms represent the constant asymptotic error bound.
\end{proof}

\section{Simulations}\label{simulations}

\color{blue}
    
In this section  we present a comparative analysis of \afl~implementations for convex objective functions for linear regression and SVM classification tasks. The experiments are conducted with synthetic non-IID data partitioning and evaluated for model performance, client delays, and convergence behavior. Key findings highlight the differing sensitivities of regression and classification models to delay and participation variability. In particular \emph{Linear Regression} which is used for predicting continuous values, while \emph{SVM} which is used for binary classification. Our main goal is to evaluate the impact of client delays, non-IID data, and client participation on \afl~for regression and classification tasks. 

Linear regression and the SVM models are defined as follows:

\paragraph{Linear Regression Model} The linear regression trained to model the relationship between input features \( x \in \mathbb{R}^d \) and a target \( y \in \mathbb{R} \) using the linear function \( f(x) = w^\top x + b \), where \( w \in \mathbb{R}^d \) is the weight vector and \( b \in \mathbb{R} \) is the bias term. The loss function is the Mean Squared Error (MSE), defined for a single sample \((x_i, y_i)\) as \[ \mathcal{L}(x_i, y_i) = \frac{1}{2} \left( y_i - (w^\top x_i + b) \right)^2. \]

\paragraph{Linear SVM Model} The SVM trained with the hinge loss for binary classification. The model's decision function is \( f(x) = w^\top x + b \), where \( w \in \mathbb{R}^d \) is the weight vector, \( b \in \mathbb{R} \) is the bias, and \( x \in \mathbb{R}^d \) is the input data. The hinge loss for a single sample \((x_i, y_i)\) is defined as
    \[ \mathcal{L}(x_i, y_i) = \max(0, 1 - y_i (w^\top x_i + b)), \]
    where \( y_i \in \{-1, 1\} \) represents the target label. The optimization objective for \( N \) samples is to minimize the empirical hinge loss, given by 
    \[ \mathcal{L} = \frac{1}{N} \sum_{i=1}^N \max(0, 1 - y_i (w^\top x_i + b)).\]

Before presenting the detailed derivations, we first clarify the key definitions and notations used in the subsequent analysis to ensure consistency and readability.

\paragraph{Data Partitioning}
Synthetic data is generated for both tasks: 

(i) \emph{Regression Data:} Continuous features sampled uniformly. We generate a synthetic dataset for linear regression with \( 2000 \) samples and \( 10 \) input features, where the input values are uniformly sampled from the range \([-5, 5]\). The weights $w$ are drawn from a uniform distribution \([2.0, 4.0]\), and the target values are computed as a linear combination of the features, with a bias term of $q=5$ and additive Gaussian noise drawn from a normal distribution with mean $0$ and standard deviation $0.2$ to simulate randomness in the measurements.

(ii) \emph{Classification Data:} Binary labels derived from logits \footnote{The term \emph{logits}, computed as the weighted sum of input features with an added bias, represents the raw decision scores before classification; in this dataset, they are thresholded at zero to assign binary labels.}. We generate a synthetic dataset for binary classification with \( 2000 \) samples and \( 10 \) input features, where the input values are uniformly sampled from the range \([-5, 5]\). The weights $w$ are drawn from a uniform distribution \([2.0, 4.0]\), and the logits are computed as a linear combination of the features with an added bias of $5$. The binary labels are assigned based on whether the logits are greater than zero, resulting in labels $0$ or $1$.

The data is partitioned non-IID across clients using a Dirichlet distribution with concentration parameter \( \zeta=0.5 \) to control heterogeneity \cite{lin2016dirichlet}. Moreover, each client trains its own local model on non-IID data by minimizing the the MSE and hinge loss using SGD for regression and binary classification, respectively.

In \afl~client updates arrive at the server at different times due to varying computation speeds, network conditions, and resource availability, resulting in update delays. To account for this, a delay factor \( \tau_t \) is introduced, representing the time difference between the slowest and fastest client updates in a given round. Formally, \( \tau_t = \max(\tau_c) - \min(\tau_c) \) for all clients \( c \) participating in the round \( t \), where \( \tau_c \) denotes the delay or execution time for client \( c \). The delay \( \tau_t \) influences the learning rate \( \gamma_t \) to prevent stale updates from disproportionately affecting the global model. Specifically, the learning rate is adjusted according to \( \gamma_t = \frac{\gamma_0}{\sqrt{t + 1} \cdot (1 + \alpha \cdot \tau_t)} \), where \( \gamma_0 \) is the initial learning rate, \( t \) is the epoch index, and \( \alpha \) is a scaling factor that controls how much \( \tau_t \) influences the adjustment. When \( \tau_t \) is small, meaning that client delays are relatively uniform, the adjustment factor remains close to 1, and the learning rate behaves similarly to its default value. However, when \( \tau_t \) is large, indicating significant discrepancies in client delays, the learning rate is reduced to account for the staleness of delayed updates and to stabilize convergence. This mechanism is crucial for ensuring that the \afl~process remains robust and fair, allowing slower clients to contribute meaningfully without destabilizing the global model. Empirical observations from the experiments indicate that regression tasks are less sensitive to \( \tau_t \) due to the smoothing effect of the quadratic loss function, while SVM classifiers, which rely on hinge loss, show a greater sensitivity to large \( \tau_t \), necessitating more careful tuning of the \( \alpha \) parameter to maintain stability.

\subsection{Implementation Details}

The implementation of \afl~is carried out in \texttt{Python}, utilizing \texttt{PyTorch} for model construction and training \cite{paszke2019pytorch}. Parallelism across client updates is managed using the \texttt{asyncio} library, which facilitated concurrent execution and asynchronous communication between clients and the server. Several supporting libraries are employed, including \texttt{NumPy} for numerical computations, \texttt{Matplotlib} for visualizing performance metrics and trends. The optimization of both the regression and SVM models is performed using SGD, ensuring an efficient and flexible approach to updating model parameters.

\paragraph{Code and Hardware}
The \afl~framework is implemented in \texttt{PyTorch}, with code available at Github \footnote{\url{https://github.com/Ali-Forootani/Asynchronous-Federated-Learning}} and archived on Zenodo \footnote{\url{ https://doi.org/10.5281/zenodo.14548841}, DOI: 10.5281/zenodo.14548841} \cite{Forootani_CADNN}. Simulations are executed on a machine with a 13\textsuperscript{th} Gen Intel\textsuperscript{\textregistered} Core\texttrademark{} i5-1335U processor (10 cores, 12 threads, 4.6 \texttt{GHz}).

\subsection{Linear Regression Model with \afl} 
In Table \ref{linear_reg} the hyperparameters of the \afl~algorithm for the linear regression are shown.

	\begin{table}[h]
    \color{blue}
		\centering
		\begin{tabular}{ll}
			\toprule
			\textbf{Parameter} & \textbf{Value} \\
			\midrule
			Number of clients (\( C \)) & 10 \\
			Total rounds (\( \mathcal{J} \)) & 400 \\
			Local epochs (\( I \)) & 50 \\
			Initial learning rate (\( \gamma_0 \)) & 0.001 \\
			Delay adjustment factor (\( \alpha \)) & 0.01 \\
			Batch size & 32 \\
			Input dimensions & 10 \\
			\bottomrule
		\end{tabular}
		\caption{\color{blue}Hyperparameters used in the experiments for the \afl~in linear regression.}
        \label{linear_reg}
	\end{table}

\begin{figure}[h]

	\centering
	\includegraphics[width=0.7\linewidth]{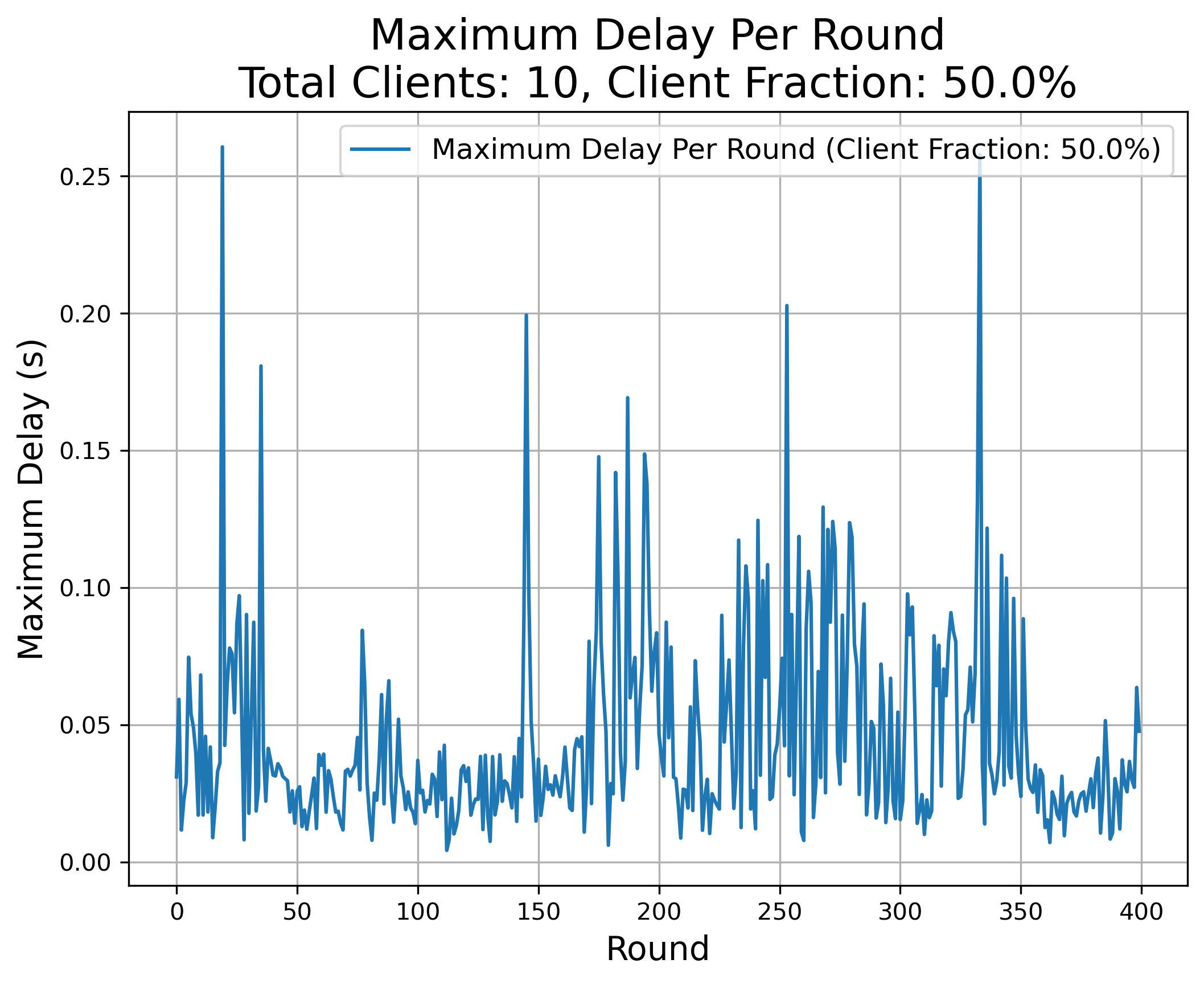}
	\caption{\color{blue}Maximum delay per round during asynchronous updates for the linear regression model.}
	\label{fig:max_delays_reg}
\end{figure}

\begin{figure}
    \centering
    \includegraphics[width=0.7\linewidth]{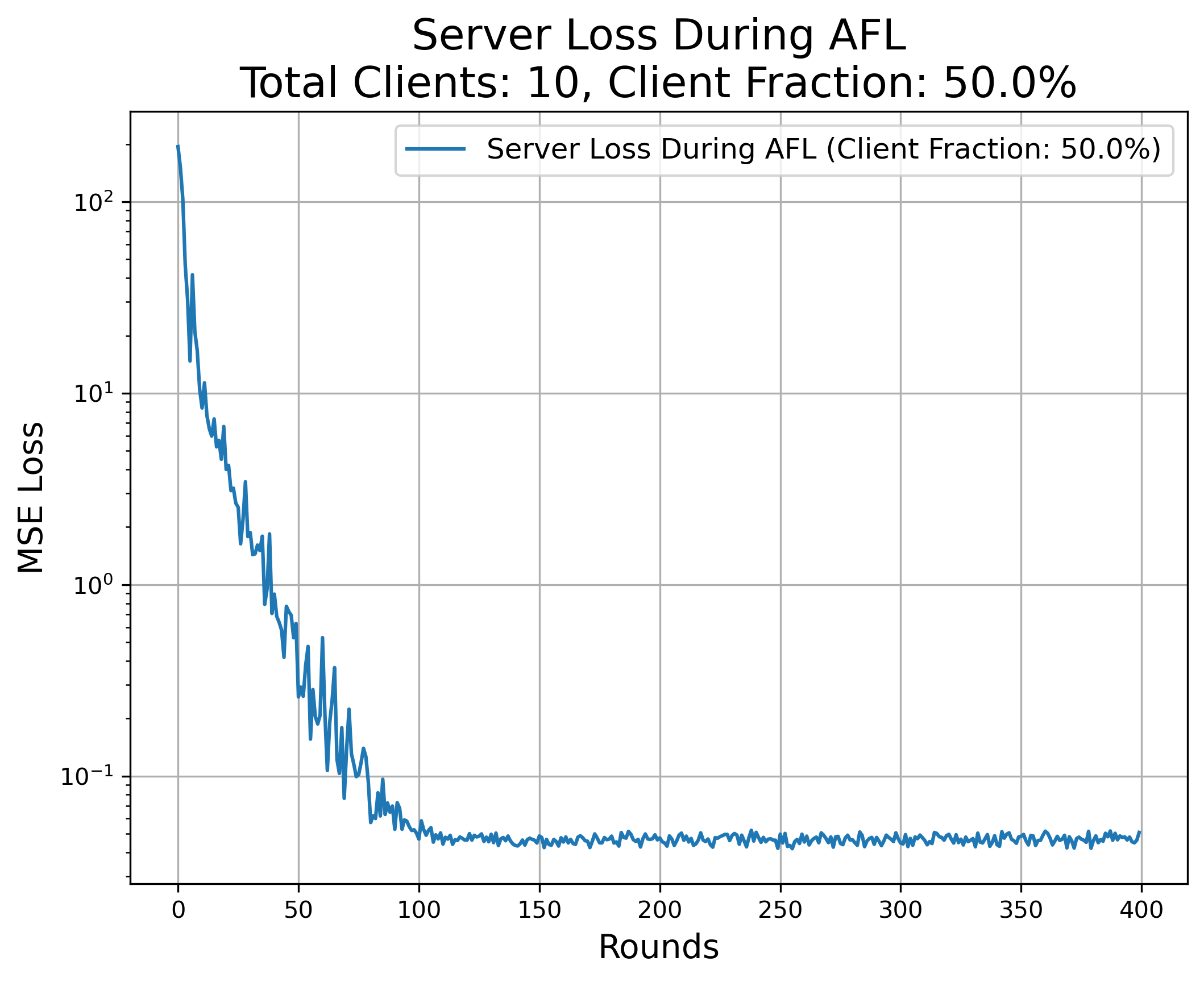}
    \caption{\color{blue}Server loss per round during asynchronous updates in the linear regression model.}
    \label{fig:server_loss_reg}
\end{figure}

In the first experiment client fraction is assumed $0.5$, i.e. \( J=5 \). The maximum delays observed per round are plotted in Figure \ref{fig:max_delays_reg} for linear regression model. As we assumed, the delay is within an acceptable threshold, meaning the clients do not drift from each other in each round \footnote{It is worth to highlight that the delay is a function of the processor and differs in another hardware.}. The server loss is depicted in Figure \ref{fig:server_loss_reg} which we see the convergence of the server within an acceptable bound. To quantify efficiency, we computed the cumulative wall-clock time as 
\(
T_{\mathrm{cum}}(k) = \sum_{t=1}^{k} \big(\max_{c \in C_t} \tau_c - \min_{c \in C_t} \tau_c\big),
\)
where $\tau_c$ denotes the execution time of client $c$ in round $t$, and introduced an energy proxy 
\(
E_{\mathrm{proxy}}(k) = \sum_{t=1}^{k} \sum_{c \in C_t} P_c \, \tau_c,
\)
with $P_c$ representing an estimated power consumption per client, enabling joint assessment of temporal and computational resource costs. For reproducibility, the power coefficients were chosen based on typical thermal design power (TDP) values reported for mid‑range GPUs ($125W$) and multicore CPUs ($45W$) in existing energy‑aware machine learning studies, rather than measured consumption on our setup \cite{henderson2020towards}.

\begin{figure}
    \centering
    \includegraphics[width=0.7\linewidth]{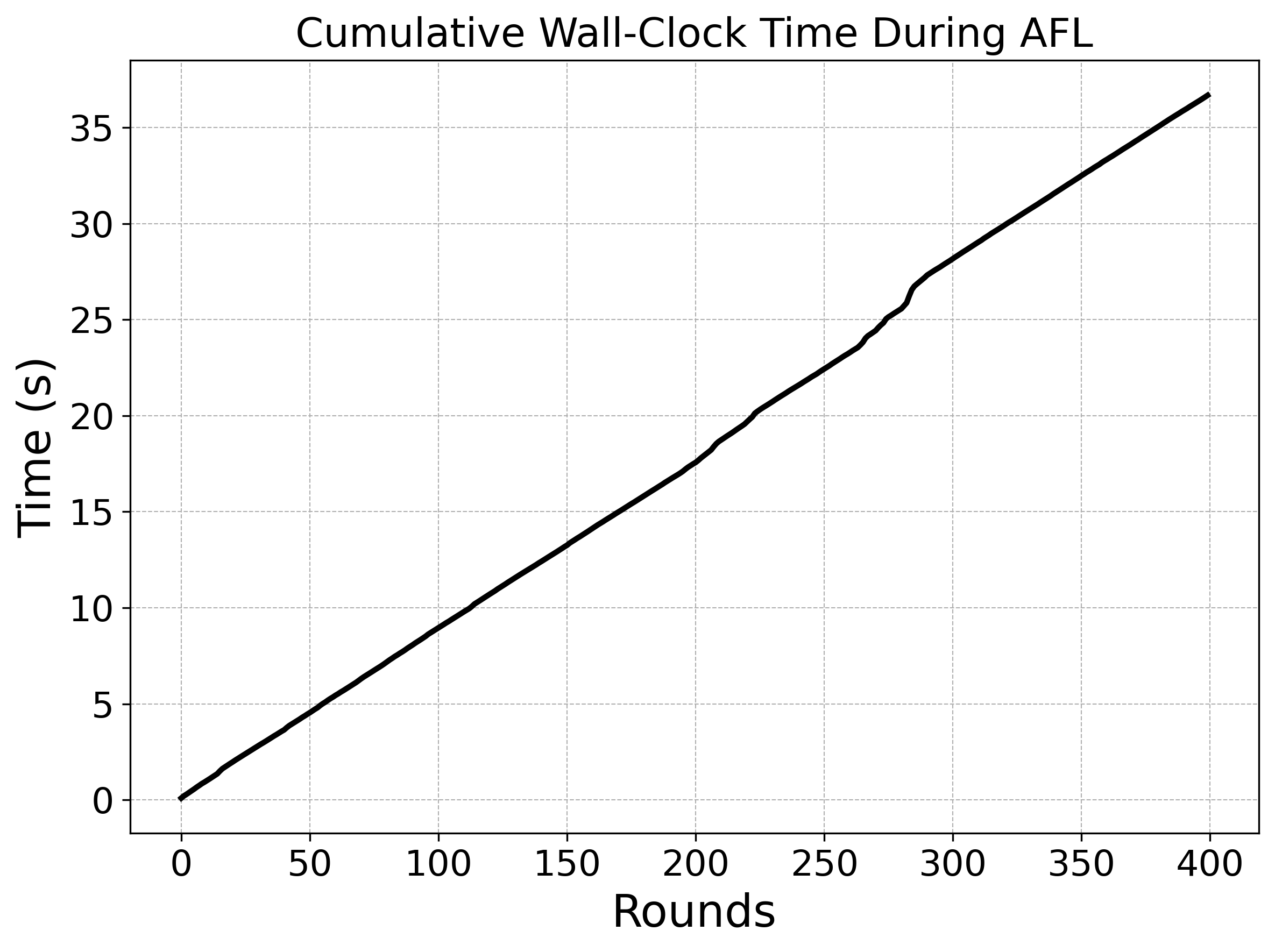}
    \caption{\color{blue}Cumulative wall-clock time during \afl~for the regression model.}
    \label{fig:wall_clock_reg}
\end{figure}

\begin{figure}
    \centering
    \includegraphics[width=0.7\linewidth]{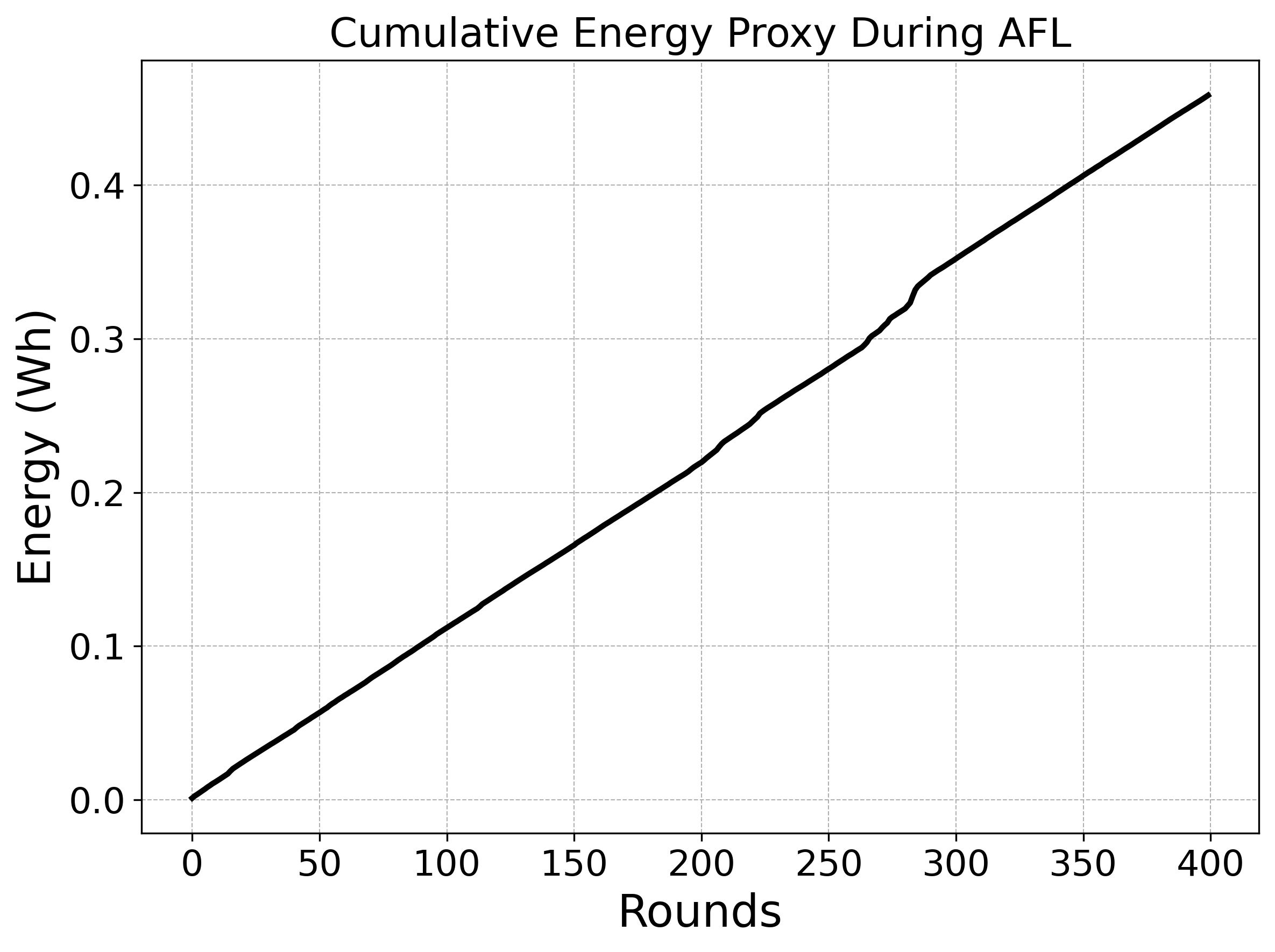}
    \caption{\color{blue}Cumulative energy proxy during \afl~for the regression model.}
    \label{fig:energy_proxy_reg}
\end{figure}

The cumulative wall‑clock time shown in Figure \ref{fig:wall_clock_reg} increases nearly linearly with the number of rounds because each asynchronous round is dominated by the slowest client's completion time, which remains relatively stable across rounds. Minor deviations (small steps or changes in slope) reflect fluctuations in client delays caused by data heterogeneity and asynchronous update timing. The energy proxy shown in Figure \ref{fig:energy_proxy_reg} grows proportionally with the wall‑clock time since it is computed as elapsed time multiplied by an estimated device‑specific power draw. Its almost linear shape indicates consistent per‑round computational load, with small slope variations capturing round‑to‑round differences in execution time across participating clients.

\begin{figure}
    \centering
    \includegraphics[width=0.8\linewidth]{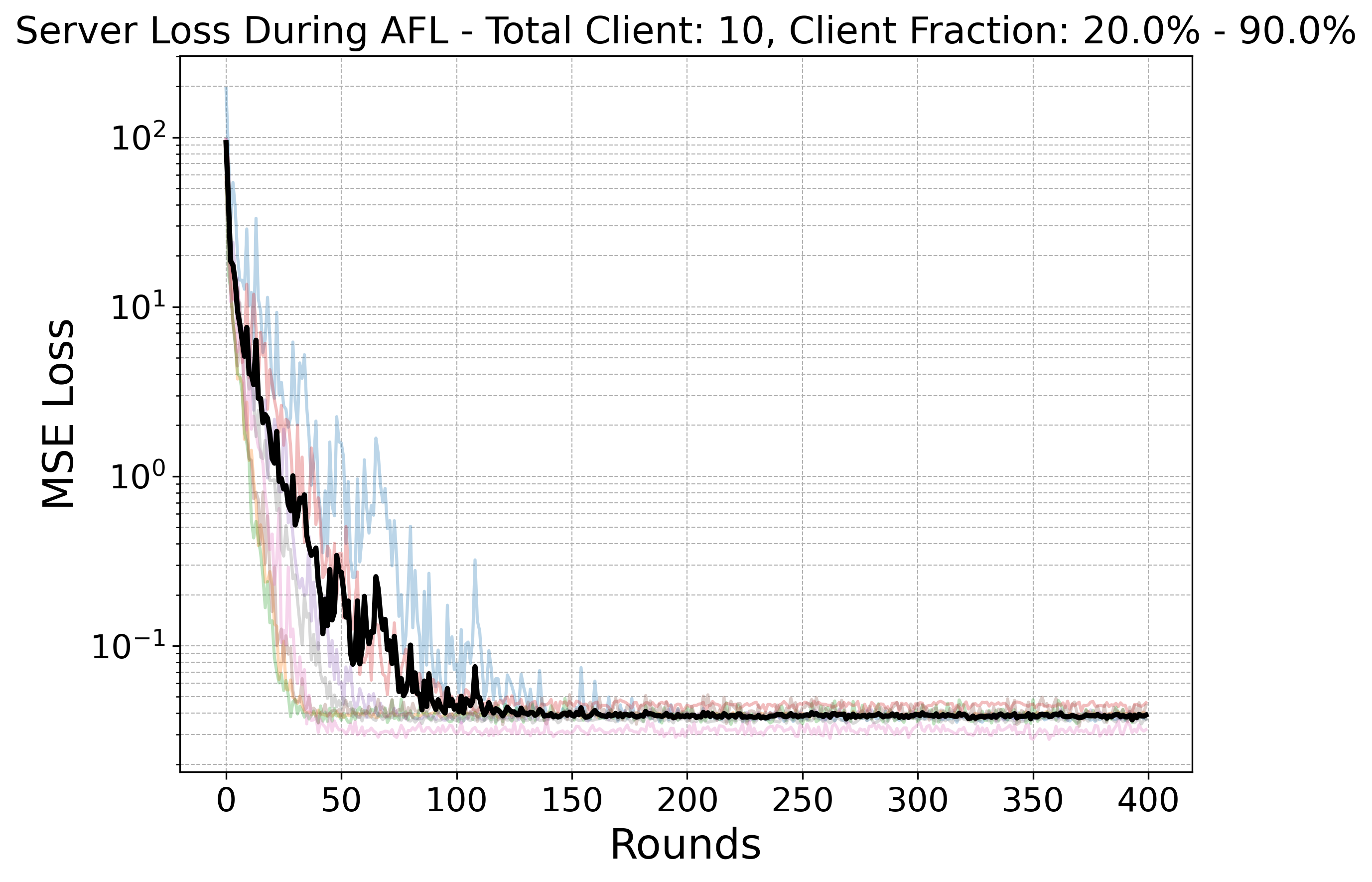}
    \caption{\color{blue}Average server loss in \afl~ when client fraction varies from $20\% - 90\%$ in regression task.}
    \label{afl_regression_plots}
\end{figure}

We conduct a series of experiments to analyze the impact of varying client participation fractions in an \afl~setting for a convex regression problem. We varied the client fraction from 0.2 (20\%) to 0.9 (90\%) to investigate the effect of different levels of client participation per round. We run multiple \afl~simulations, each corresponding to a specific client fraction. After each training round, we recorded the server loss (Mean Squared Error) and stored the results. Loss curves for all client fraction settings are plotted on  Figure \ref{afl_regression_plots}. The average loss curve is computed across all runs to observe overall trends in model convergence. The server loss consistently decreased across all experiments, confirming model convergence. Higher client fractions (e.g., 90\%) results in faster convergence due to the larger number of updates received per round. Lower client fractions (e.g., 20\%) exhibits slower convergence and higher variance in loss values due to reduced client contributions in each round.

\begin{figure}
    \centering
    \includegraphics[width=0.8\linewidth]{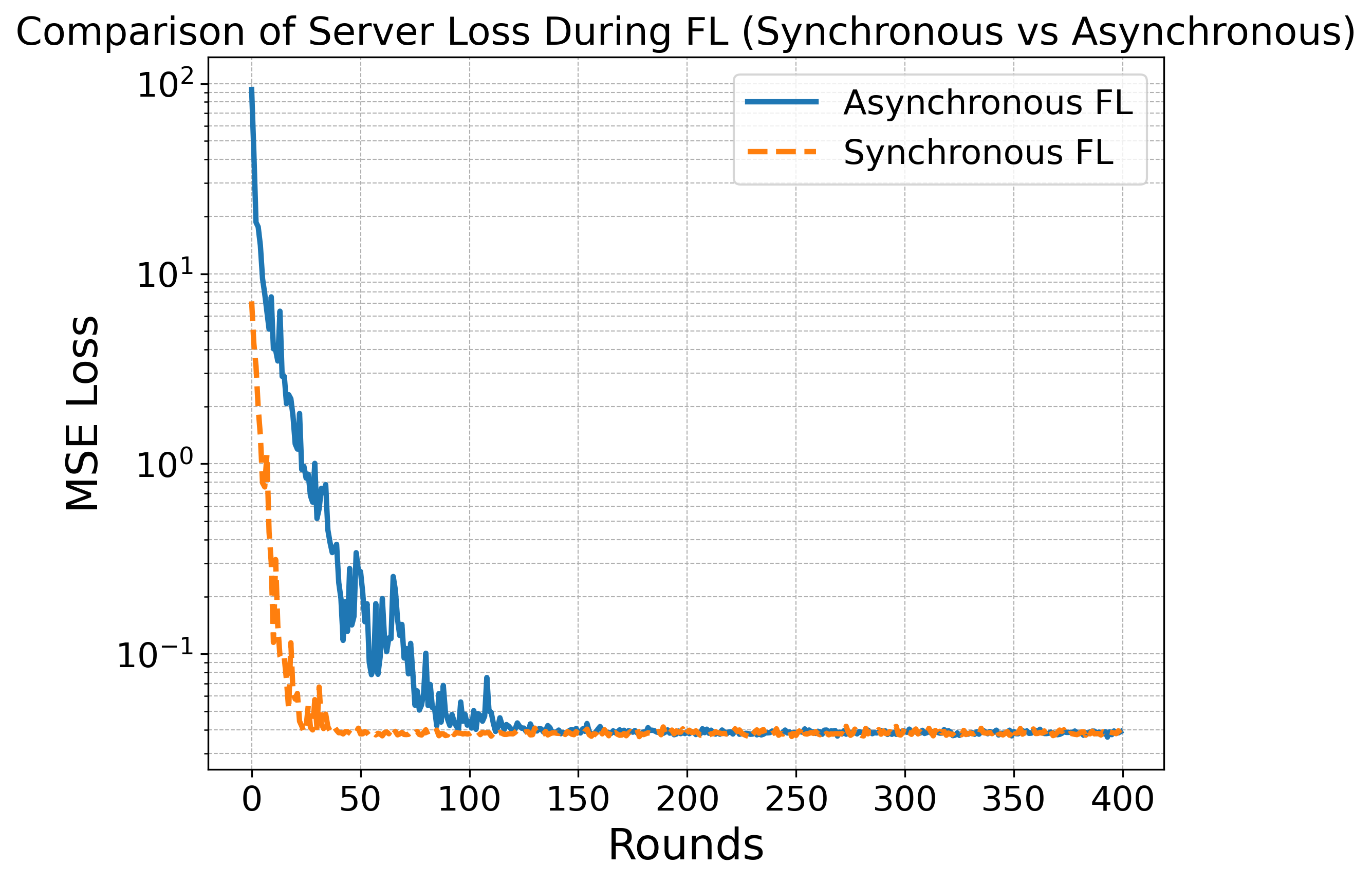}
    \caption{\color{blue}Comparing the average of server loss trends between Synchronous \fl~and \afl~ when client fraction varies from $20\% - 90\%$ in regression task.}
    \label{comparison_afl_sfl_regression}
\end{figure}

We compare the server loss trends between Synchronous \fl~(\syncfl) and \afl~over $400$ rounds. For both setups, client fractions were varied between 0.2 (20\%) and 0.9 (90\%), and the average loss curves are plotted in Figure \ref{comparison_afl_sfl_regression}. As we expected, \syncfl~demonstrates faster convergence during the early rounds compared to \afl, achieves a lower loss value more quickly, stabilizing earlier around \(10^{-2}\). The consistent curve indicates a more stable training process, due to synchronous updates where all participating clients contribute at the same time. \afl    (blue solid line) converges more slowly in the early rounds, exhibiting higher loss variability due to asynchronous updates with variable delays. It achieves comparable final loss values after approximately $200$ rounds, matching the performance of \syncfl.

\subsection{Linear SVM Classification with \afl}

In Table \ref{svm_classification} the hyperparameters of the \afl~algorithm for the SVM binary classification are shown. 
    
	\begin{table}[h]
    \color{blue}
		\centering
		\begin{tabular}{ll}
			\toprule
			\textbf{Parameter} & \textbf{Value} \\
			\midrule
			Number of clients (\( C \)) & 10 \\
			Total rounds (\( \mathcal{J} \)) & 1000 \\
			Local epochs (\( I \)) & 100 \\
			Initial learning rate (\( \gamma_0 \)) & 0.0005 \\
			Delay adjustment factor (\( \alpha \)) & 0.01 \\
			Batch size & 32 \\
			Input dimensions & 10 \\
			\bottomrule
		\end{tabular}
		\caption{\color{blue}Hyperparameters used in the experiments for the \afl~in linear SVM for classification.}
        \label{svm_classification}
	\end{table}

The SVM classifier showed greater sensitivity to client delays due to the nature of hinge loss. 

\begin{figure}[h]
	\centering
	\includegraphics[width=0.7\linewidth]{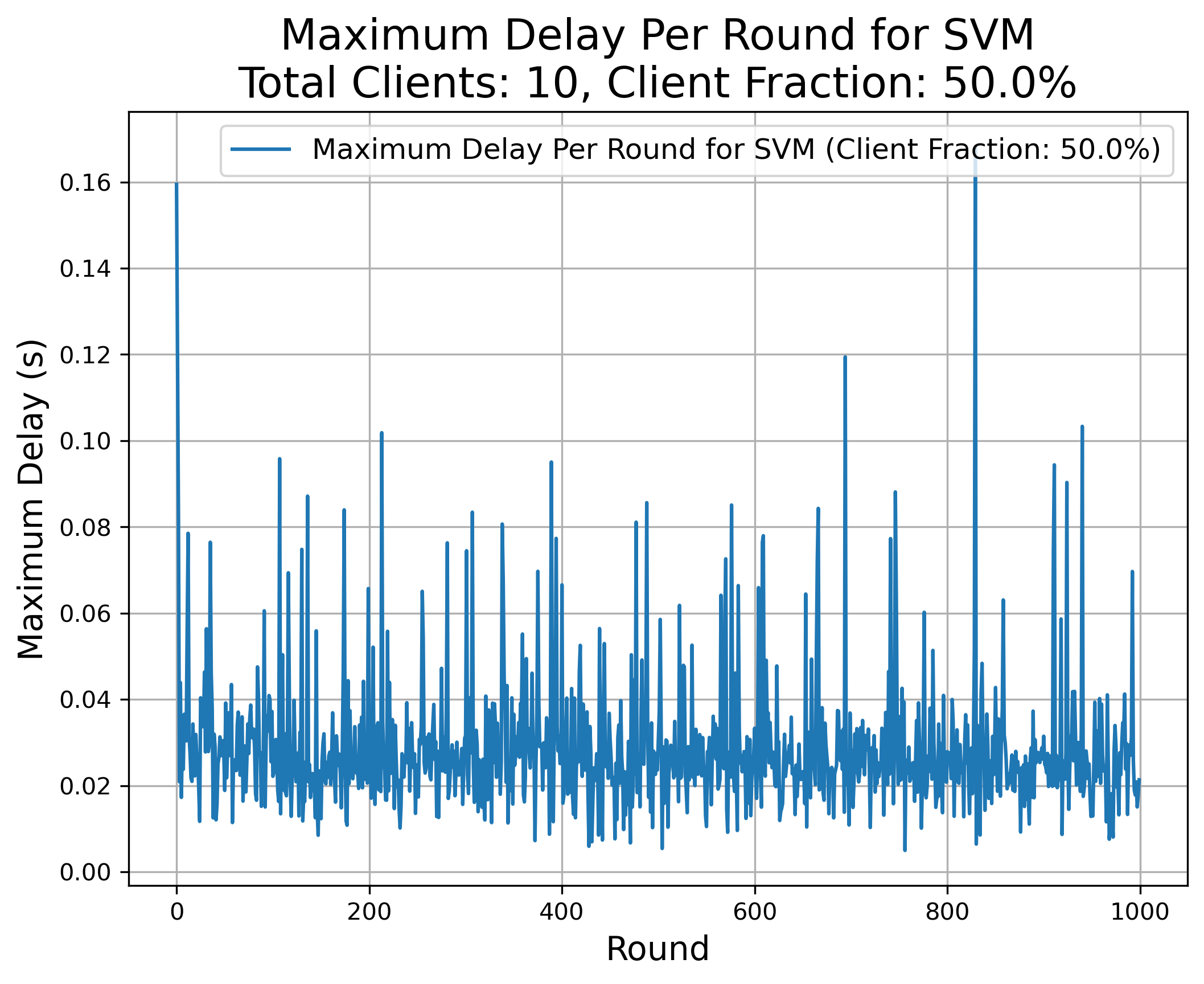}
	\caption{\color{blue}Maximum delay per round during asynchronous updates for the linear SVM model with client fraction $50\%$.}
	\label{fig:max_delays}
\end{figure}

\begin{figure}[h]
	\centering
	\includegraphics[width=0.7\linewidth]{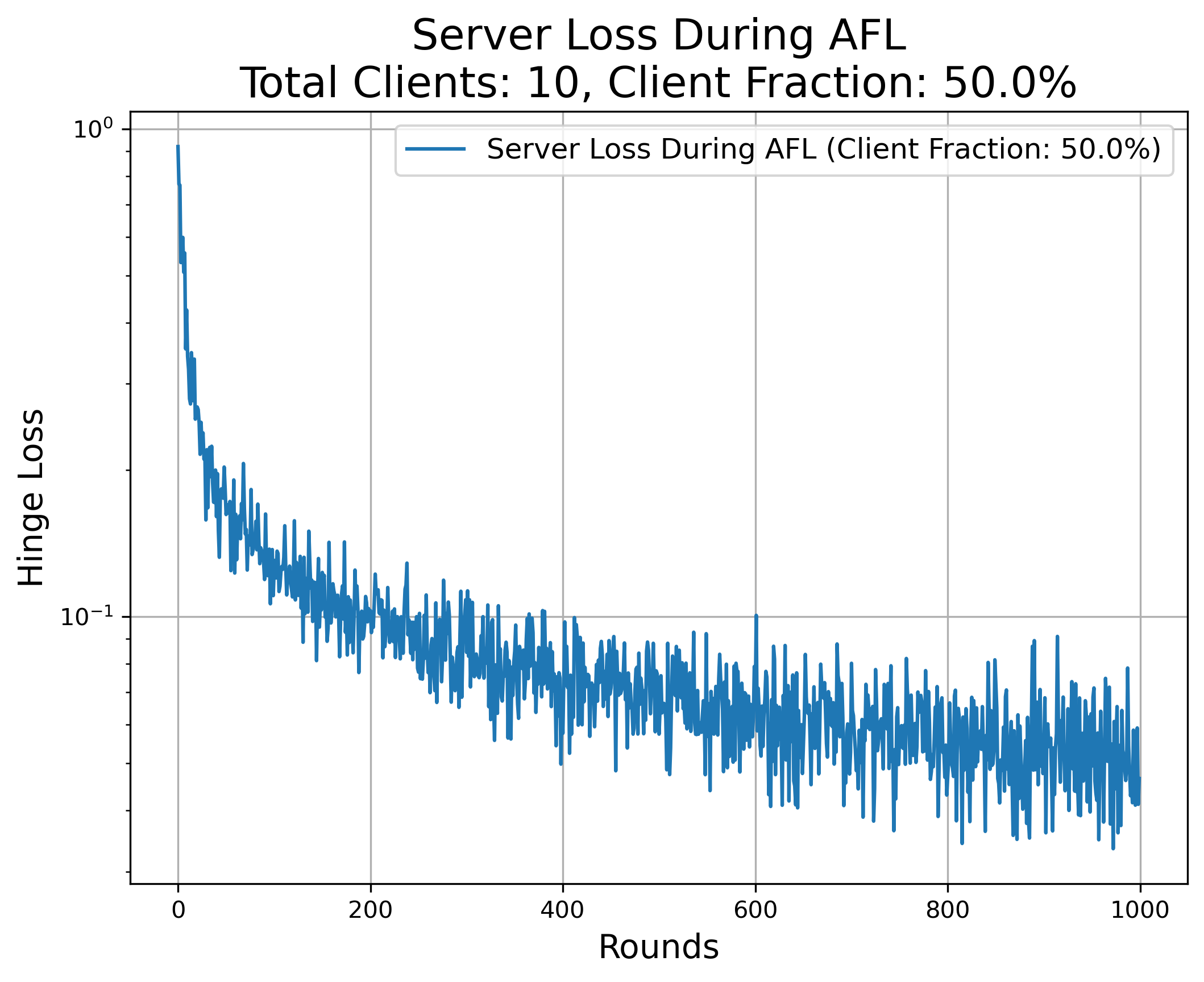}
	\caption{\color{blue}Server loss per round during asynchronous updates for the linear SVM model with client fraction $50\%$.}
	\label{fig:max_delays}
\end{figure}

The plot illustrates the server hinge loss over $1000$ rounds of \afl~with linear SVM for binary classification, where $50\%$ of the clients (5 out of 10) participate in each round. The loss decreases sharply in the initial rounds, indicating rapid learning, and stabilizes after around $200–300$ rounds. However, fluctuations persist throughout the later stages, likely due to the reduced diversity in client updates caused by the lower client fraction and the non-IID nature of the data. 

The model converges rapidly in the early rounds, with a significant drop in hinge loss, indicating fast learning due to the convex nature of the problem. However, after approximately $200–300$ rounds, the loss stabilizes but shows slight fluctuations, likely due to asynchronous updates and the non-IID distribution of client data.

By analyzing the server loss across 1000 training rounds with client fractions ranging from 20\% to 90\%, it was observed that higher client fractions contribute to faster and more stable convergence due to the increased frequency and diversity of updates. Lower client fractions, on the other hand, exhibit higher variability and slower loss reduction, as fewer clients participate in each round, leading to greater stochasticity in updates. The average loss curve, represented by the bold black line, summarizes the overall trend, showing a sharp decline in the early rounds and a slower, more stable reduction over time. However, fluctuations in loss values become more apparent in later rounds, likely due to the combined effects of asynchronous updates and non-IID data distributions across clients. The findings underscore the trade-off between asynchrony and client participation, where larger client fractions mitigate variability but require more computational resources. Overall, this analysis highlights the importance of balancing client fraction and asynchrony to optimize the performance of \afl~systems.

\begin{figure}
    \centering
    \includegraphics[width=0.9\linewidth]{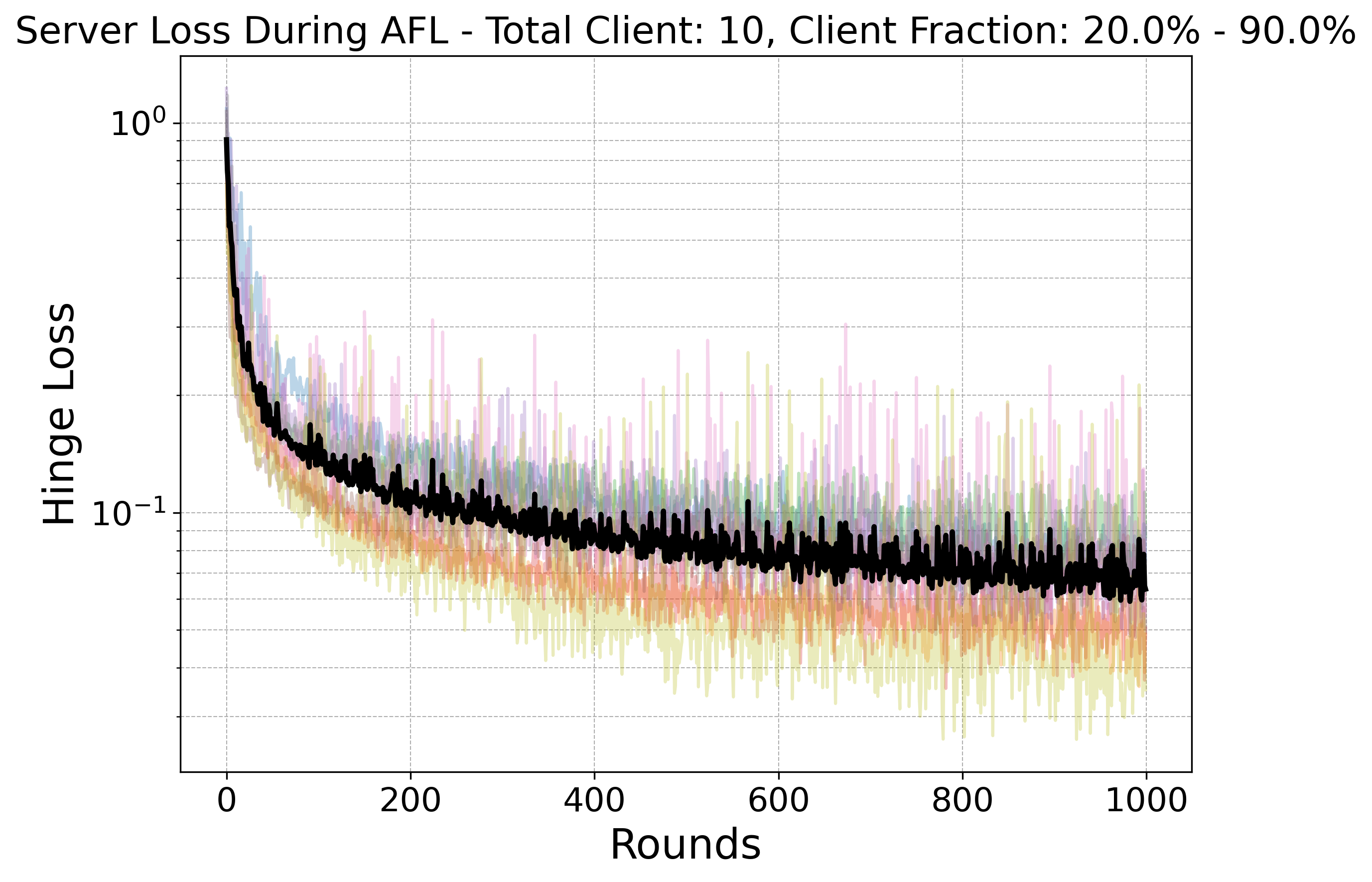}
    \caption{\color{blue}Average of the server loss in \afl~algorithm for SVM based classification task when client fraction varies between $20\%$ - $90\%$.}
    \label{fig:enter-label}
\end{figure}

The comparison between \texttt{SFL} and \afl~highlights key differences in convergence dynamics and stability. \texttt{SFL}, represented by the blue curve, consistently achieves faster and more stable convergence with lower server loss due to synchronized updates from all clients after each round. In contrast, \afl, shown by the orange dashed curve, converges more slowly and stabilizes at a higher loss due to the inherent variability caused by asynchronous updates and client delays. The variability in \afl~is more pronounced, particularly in later rounds, reflecting the challenges of aggregating non-uniform updates. While \texttt{SFL} benefits significantly from higher client fractions, ensuring smooth and robust loss reduction, \afl~offers flexibility and scalability at the cost of increased variability and reduced convergence efficiency. This trade-off underscores the need for hybrid strategies to balance the advantages of both approaches in practical deployments.

\begin{figure}
    \centering
    \includegraphics[width=0.9\linewidth]{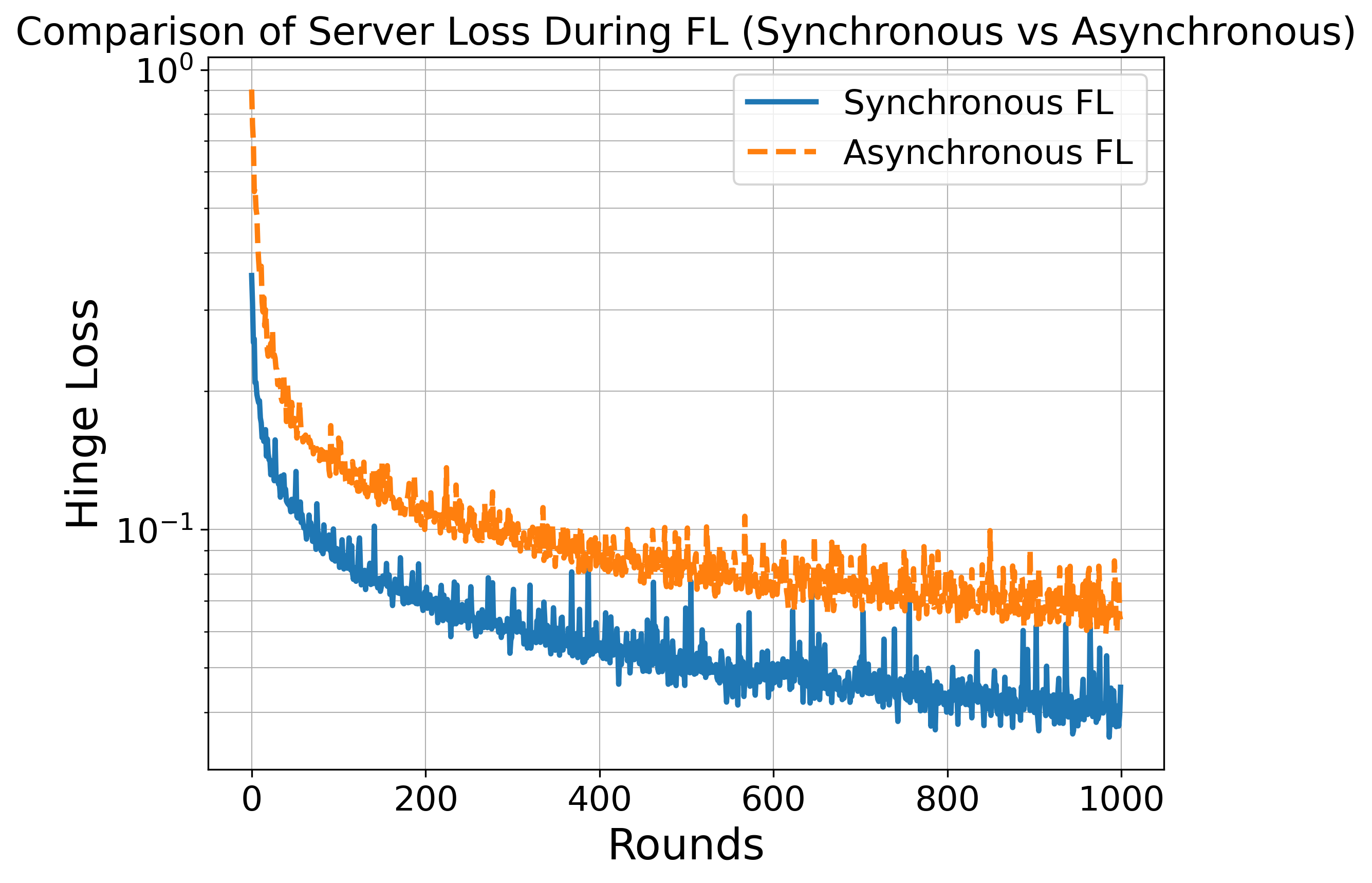}
    \caption{\color{blue}Comparing the average of server loss trends between Synchronous \fl~and \afl~ when client fraction varies from $20\% - 90\%$ in SVM classification task.}
    \label{fig:enter-label}
\end{figure}

In addition to evaluating the aggregated global model, we note that the performance of local models on their respective local datasets aligns with their contributions to the federated aggregation. Future work will include a detailed reporting of local model performance to provide a more comprehensive analysis of asynchronous federated learning behavior.

\color{black}

\section{Conclusion}\label{conclusion}

In this paper we proposed \afl~algorithm that significantly improved the scalability, efficiency, and robustness of federated learning systems, especially in scenarios involving heterogeneous client populations and dynamic network conditions. By allowing clients to update the global model asynchronously, \afl~mitigated the inefficiencies and delays associated with synchronous methods, leading to faster convergence and better resource utilization. Our theoretical analysis, supported by variance bounds and martingale difference sequence theory, demonstrates that \afl~can effectively handle client delays and model staleness, ensuring reliable convergence even in challenging environments. \color{blue}The practical applicability of \afl~is further validated through the successful training of a decentralized linear regression model and SVM classifications, highlighting its potential for large-scale, privacy-preserving applications in diverse, real-world settings. This work paves the way for more efficient distributed learning frameworks, particularly in resource-constrained and dynamically changing environments.\color{black}

The algorithm successfully leverages an asynchronous federated learning framework to enable multiple clients to train local models independently while collaborating through a central server. Key features such as dynamic learning rate adjustment, delay-aware optimization, weighted aggregation, and early stopping ensure robust and efficient training across heterogeneous clients with diverse data distributions.

The results highlighted the adaptability of the \afl~algorithm to various configurations, as evidenced by its ability to reduce training loss over iterations and maintain fairness in client participation. Scenario-based evaluations confirm the algorithm's scalability and capability to address real-world challenges such as non-uniform client contributions and variable training conditions.


\color{blue} While our analysis focuses on convergence under bounded delays and convex client objectives, several critical aspects remain open for future research. Specifically, \afl~is inherently more vulnerable to Byzantine attacks due to its asynchronous nature, which motivates integrating robust aggregation rules. Moreover, the communication cost in large-scale deployments can be reduced by adopting compression and event-triggered update schemes. Finally, fairness guarantees under heterogeneous client delays require adaptive re-weighting or client selection mechanisms. Addressing these aspects, in particular the extension to non-convex objectives already explored in our ongoing work\footnote{See e.g. \texttt{Github} \url{https://github.com/Ali-Forootani/AFL_non_convex_non_iid} or \texttt{Zenodo} \url{https://zenodo.org/records/14962410}.}, constitutes a promising direction for future research.
\color{black}


\appendix

To simplify \eqref{last_eq_simple_random_sampling} the expression, we define:
\[
 \Gamma = \sum_{c=1}^J \sum_{i=0}^{I-1} \E\left[\norm{\sum_{i=1}^{c} \sum_{j=0}^{p_{c,i}(i)} \left(\rvx_{\psi_i} - \overline{\vx}\right)}^2\right].
\]

We start by expanding \(\Gamma\):
\begin{multline*}
 \Gamma  = \frac{CI^3\nu^2}{C-1} \sum_{c=1}^J (c-1) - \frac{I^3\nu^2}{C-1} \sum_{c=1}^J (c-1)^2 \\
 + J\nu^2 \sum_{i=0}^{I-1} i^2 - \frac{2I\nu^2}{C-1} \sum_{c=1}^J (c-1) \sum_{i=0}^{I-1} i.
\end{multline*}

Substituting the known summation formulas, we have:
\begin{multline*}
 \Gamma  = \frac{CI^3}{C-1} \frac{(J-1)J}{2} - \frac{I^3}{J-1} \frac{(J-1)J(2J-1)}{6} \\
 + J \frac{(I-1)I(2I-1)}{6} - \frac{2I}{J-1} \frac{(J-1)J}{2} \frac{(I-1)I}{2}.
\end{multline*}

Next, we separate the first term into two components:
\begin{multline*}
 \Gamma  = \frac{(J-1)JI^3}{2} + \frac{I^3}{C-1} \frac{(J-1)J}{2} \\ - \frac{I^3}{C-1} \frac{(J-1)J(2J-1)}{6} \\
 + C \frac{(I-1)I(2I-1)}{6} - \frac{2I}{C-1} \frac{(J-1)J}{2} \frac{(I-1)I}{2}.
\end{multline*}

Next, we group the terms containing \(\frac{1}{C-1}\):
\begin{multline}\label{middle_relation}
 \Gamma  = \frac{(J-1)JI^3}{2} + J \frac{(I-1)I(2I-1)}{6} + \frac{I^3}{C-1} \frac{(J-1)J}{2} \\
 - \frac{I^3}{C-1} \frac{(J-1)J(2J-1)}{6} - \frac{2I}{C-1} \frac{(J-1)J}{2} \frac{(I-1)I}{2}.
\end{multline}

Now, we analyze the first two terms in \eqref{middle_relation}:
\begin{multline*}
\text{Term}_{1} + \text{Term}_{2} = \frac{J^2I^3}{2} - JI \left( \frac{3I^2}{6} - \frac{(I-1)(2I-1)}{6} \right) \\
= \frac{J^2I^3}{2} - \frac{JI(I^2 + 3I - 1)}{6} = \frac{1}{2} JI^2 (JI - 1) - \frac{1}{6} JI (I^2 - 1).
\end{multline*}

Next, we consider the remaining three terms in \eqref{middle_relation}:
\begin{multline}\label{second_middle_relation}
\text{Term}_{3} + \text{Term}_4+ \text{Term}_5 = \frac{I^2}{C-1} \Big( \frac{(J-1)JI}{2} \\ - \frac{(J-1)J(2J-1)I}{6} - \frac{(J-1)J(I-1)}{2} \Big).
\end{multline}
Rearranging \eqref{second_middle_relation} gives:
\begin{multline*}
\text{Term}_{3} + \text{Term}_4+ \text{Term}_5\\ = -\frac{I^2}{C-1} \left( \frac{(J-1)J(2J-1)I}{6} - \frac{(J-1)J}{2} \right) \\
= -\frac{(J-1)JI^2}{2(C-1)} \left( \frac{1}{3}(2J-1)I - 1 \right).
\end{multline*}

Finally, we can summarize our results to obtain:
\begin{multline}\label{third_middle_relation}
 \Gamma  = \frac{1}{2} JI^2 (JI - 1) - \frac{1}{6} JI (I^2 - 1)
 \\- \frac{1}{2} \frac{(J-1)J}{(C-1)} I^2 \left( \frac{1}{3}(2J-1)I - 1 \right),
\end{multline}
comparing \eqref{third_middle_relation} with the right hand side of \eqref{last_eq_simple_random_sampling} the results .

\bibliographystyle{IEEEtran}
\bibliography{my_refs_lcss,refs}

\end{document}

%% file: math_macros.tex
\usepackage{amsmath,amsfonts,bm, amssymb}

\usepackage{amsthm}
\usepackage{mathtools}

\def\1{\bm{1}}

\def\rve{{\mathbf{e}}}

\def\rvg{{\mathbf{g}}}

\def\rvq{{\mathbf{q}}}

\def\rvx{{\mathbf{x}}}
\def\rvy{{\mathbf{y}}}

\def\vx{{\bm{x}}}
\def\vy{{\bm{y}}}
\def\vz{{\bm{z}}}

\DeclareMathAlphabet{\mathsfit}{\encodingdefault}{\sfdefault}{m}{sl}
\SetMathAlphabet{\mathsfit}{bold}{\encodingdefault}{\sfdefault}{bx}{n}

\def\gB{{\mathcal{B}}}

\def\gD{{\mathcal{D}}}

\newcommand{\E}{\mathbb{E}}

\newcommand{\Cov}{\mathrm{Cov}}

\newcommand\inp[2]{\left\langle #1, #2 \right\rangle} 
\newcommand{\norm}[1]{\left\lVert#1\right\rVert} 
\newcommand{\Norm}[1]{\lVert#1\rVert} 

\newtheorem{assumption}{Assumption}

\newtheorem{theorem}{Theorem}

\newtheorem{lemma}{Lemma}
